\documentclass[final]{cvpr}

\usepackage{times}
\usepackage{epsfig}
\usepackage{graphicx}
\usepackage{amsmath}
\usepackage{amssymb}
\usepackage{booktabs}
\usepackage{xcolor}
\usepackage{multirow}
\usepackage{float}
\usepackage{amsfonts}
\usepackage{comment}
\usepackage{url}
\usepackage[caption=false]{subfig}
\usepackage{wrapfig}
\newcommand{\parsection}[1]{\vspace{1mm}\noindent\textbf{#1:}~}

\usepackage[pagebackref=true,breaklinks=true,colorlinks,bookmarks=false]{hyperref}

\def\cvprPaperID{****} 
\def\confYear{CVPR 2021}

\newcommand{\reals}{\mathbb{R}}
\newcommand{\tp}{^\text{T}}

\begin{document}

\title{Learning Accurate Dense Correspondences and When to Trust Them}

\author{Prune Truong \qquad Martin Danelljan \qquad Luc Van Gool \qquad Radu Timofte\\
Computer Vision Lab, ETH Zurich, Switzerland\\
\small{\texttt{\{prune.truong, martin.danelljan, vangool, radu.timofte\}@vision.ee.ethz.ch}} \\
}

\maketitle

\begin{abstract}

Establishing dense correspondences between a pair of images is an important and general problem. However, dense flow estimation is often inaccurate in the case of large displacements or homogeneous regions. For most applications and down-stream tasks, such as pose estimation, image manipulation, or 3D reconstruction, it is crucial to know \emph{when and where} to trust the estimated matches. 

In this work, we aim to estimate a dense flow field relating two images, coupled with a robust pixel-wise confidence map indicating the reliability and accuracy of the prediction. We develop a flexible probabilistic approach that jointly learns the flow prediction and its uncertainty. In particular, we parametrize the predictive distribution as a constrained mixture model, ensuring better modelling of both accurate flow predictions and outliers. Moreover, we develop an architecture and training strategy tailored for \emph{robust and generalizable} uncertainty prediction in the context of self-supervised training. Our approach obtains state-of-the-art results on multiple challenging geometric matching and optical flow datasets. We further validate the usefulness of our probabilistic confidence estimation for the task of pose estimation. Code and models are available at \url{https://github.com/PruneTruong/PDCNet}. 
\end{abstract}

\section{Introduction}

\begin{figure}[t]
\centering%
\newcommand{\wid}{0.49\columnwidth}%
\subfloat[Query image\label{fig:intro-query}]{\includegraphics[width=\wid]{ 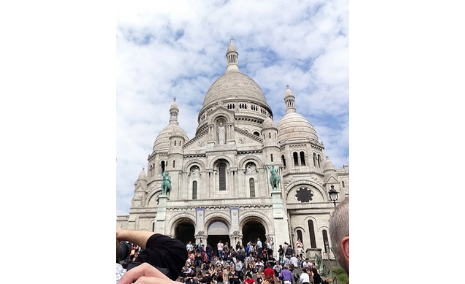}}~%
\subfloat[Reference image\label{fig:intro-ref}]{\includegraphics[width=\wid]{ 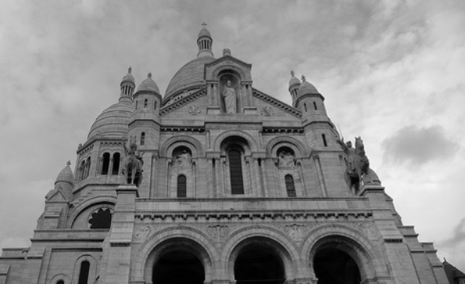}}\vspace{-2mm}
\subfloat[\centering Baseline\label{fig:intro-baseline}]{\includegraphics[width=\wid]{ 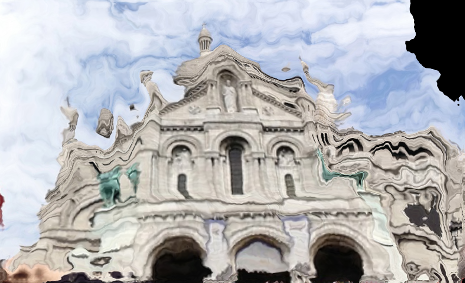}}~%
\subfloat[\centering \textbf{PDC-Net} (Ours)\label{fig:intro-PDCNet}]{\includegraphics[width=\wid]{ 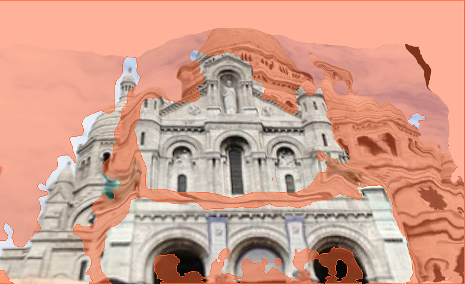}}%
\vspace{1mm}\caption{
Estimating dense correspondences between the query \protect\subref{fig:intro-query} and the reference \protect\subref{fig:intro-ref} image. The query is warped according to the resulting flows \protect\subref{fig:intro-baseline}-\protect\subref{fig:intro-PDCNet}.
The baseline \protect\subref{fig:intro-baseline} does not estimate an uncertainty map and is therefore unable to filter the inaccurate flows at \eg occluded and homogeneous regions. In contrast, our PDC-Net \protect\subref{fig:intro-PDCNet} not only estimates accurate correspondences, but also \emph{when to trust them}. It predicts a robust uncertainty map that identifies accurate matches and excludes incorrect and unmatched pixels (red).
}\vspace{-5mm}
\label{fig:intro}
\end{figure}

Finding pixel-wise correspondences between pairs of images is a fundamental computer vision problem
with numerous important applications, including dense 3D reconstruction~\cite{SchonbergerF16}, video analysis~\cite{abs-2010-04367, SimonyanZ14}, image registration \cite{shrivastava-sa11,GLAMpoint}, image manipulation~\cite{HaCohenSGL11, LiuYT11}, and texture or style transfer~\cite{Kim2019, Liao2017}.
Dense correspondence estimation has most commonly been addressed in the context of optical flow~\cite{Baker2011, Horn1981, Hur2020OpticalFE,RAFT}, where the image pairs represent consecutive frames in a video. While these methods excel in the case of small appearance changes and limited displacements, they cannot cope with the challenges posed by the more general geometric matching task. In geometric matching, the images can stem from radically different views of the same scene, often captured by different cameras and at different occasions.
This leads to large displacements and significant appearance transformations between the frames.

In contrast to optical flow, the more general dense correspondence problem has received much less attention~\cite{Melekhov2019,Rocco2018b,RANSAC-flow, GLUNet}. 
Dense flow estimation is prone to errors in the presence of large displacements, appearance changes, or homogeneous regions. It is also ill-defined in case of occlusions or in \eg sky, where predictions are bound to be inaccurate (Fig.~\ref{fig:intro-baseline}). For geometric matching applications, it is thus crucial to know \emph{when and where to trust} the estimated correspondences. For instance, pose estimation, 3D reconstruction, and image-based localization require a set of highly robust and accurate matches as input. The predicted dense flow field must therefore be paired with a \emph{robust} confidence estimate (Fig.~\ref{fig:intro-PDCNet}).
Uncertainty estimation is also indispensable for safety-critical tasks, such as autonomous driving and medical imaging. In this work, we set out to expand the application domain of dense correspondence estimation by learning to predict reliable confidence values.

We propose the Probabilistic Dense Correspondence Network (PDC-Net), for joint learning of dense flow and uncertainty estimation, applicable even for extreme appearance and view-point changes. Our model predicts the conditional probability density of the flow, parametrized as a constrained mixture model.
However, learning \emph{reliable and generalizable} uncertainties without densely annotated real-world training data is a highly challenging problem. Standard self-supervised techniques \cite{Melekhov2019, Rocco2017a,GLUNet} do not faithfully model real motion patterns, appearance changes, and occlusions.
We tackle this challenge by introducing a carefully designed architecture and improved self-supervision to ensure robust and generalizable uncertainty predictions.

\newcommand{\bp}[1]{\textbf{#1}}

\begin{figure}[t]
\centering%
\includegraphics*[width=\columnwidth, trim=0 0 850 600]{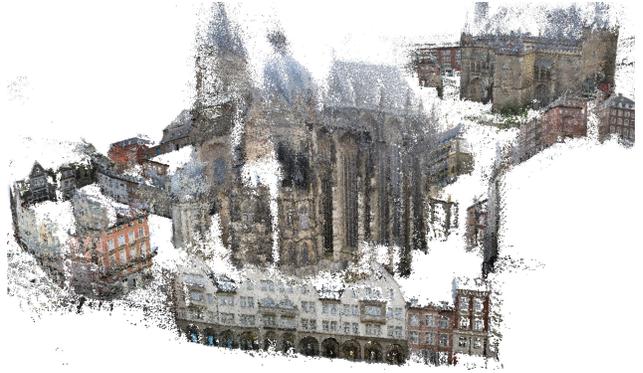}
\caption{3D reconstruction of Aachen \cite{SattlerMTTHSSOP18} using the dense correspondences and uncertainties predicted by PDC-Net.}\vspace{-5mm} 
\label{fig:aachen}
\end{figure}

\parsection{Contributions} 
Our main contributions are as follows. 
\bp{(i)}  We introduce a \emph{constrained mixture model} of the predictive distribution, allowing the network to flexibly model both accurate predictions and outliers with large errors.
\bp{(ii)} We propose an architecture for predicting the parameters of our predictive distribution, that carefully exploits the information encoded in the correlation volume, to achieve generalizable uncertainties.
\bp{(iii)} We improve upon self-supervised data generation pipelines to ensure more robust uncertainty estimation.  
\bp{(iv)} We utilize our uncertainty measure to address extreme view-point changes by iteratively refining the prediction.
\bp{(v)} We perform extensive experiments on a variety of datasets and tasks. In particular, our approach sets a new state-of-the-art on the Megadepth geometric matching dataset~\cite{megadepth}, on the KITTI-2015 training set~\cite{Geiger2013}, and outperforms previous dense methods for pose estimation on the YFCC100M dataset~\cite{YFCC}. Moreover, without further post-processing, our confident dense matches can be directly input to 3D reconstruction pipelines~\cite{SchonbergerF16}, as shown in Fig.~\ref{fig:aachen}. 

\section{Related work}

\parsection{Confidence estimation in geometric matching}
Only very few works have explored confidence estimation in the context of dense geometric or semantic matching. 
Novotny~\etal~\cite{NovotnyCVPR18Self} estimate the reliability of their trained descriptors by using a self-supervised probabilistic matching loss for the task of semantic matching.
A few approaches~\cite{DCCNet, Rocco20, Rocco2018b} represent the final correspondences as a 4D correspondence volume, thus inherently encoding a confidence score for each tentative match. However, these approaches are usually restricted to low-resolution images, thus hindering accuracy. Moreover, generating one final confidence value for each match is highly non-trivial since multiple high-scoring alternatives often co-occur.
Similarly, Wiles~\etal~\cite{D2D} learn dense descriptors conditioned on an image pair, along with their distinctiveness score. However, the latter is trained with hand-crafted heuristics, while we instead do not make assumption on what the confidence score should be, and learn it directly from the data with a single unified loss. 
In DGC-Net, Melekhov~\etal~\cite{Melekhov2019} predict both dense correspondence and matchability maps relating image pairs. However, their matchability map is only trained to identify out-of-view pixels rather than to reflect the actual reliability of the matches. 
Recently, Shen~\etal~\cite{RANSAC-flow} proposed RANSAC-Flow, a two-stage image alignment method, which also outputs a matchability map. It performs coarse alignment with multiple homographies using RANSAC on off-the-shelf deep features, followed by fine alignment.
In contrast, we propose a unified network that estimates probabilistic uncertainties. 

\parsection{Uncertainty estimation in optical flow} 
While optical-flow has been a long-standing subject of active research, only a handful of methods provide uncertainty estimates. 
A few approaches~\cite{Aodha2013LearningAC, Barron94performanceof, KondermannKJG07, KondermannMG08, KybicN11} treat the uncertainty estimation as a post-processing step. 
Recently, some works propose probabilistic frameworks for joint optical flow and uncertainty prediction instead. They either estimate the model uncertainty~\cite{ GalG15, IlgCGKMHB18}, termed epistemic uncertainty~\cite{KendallG17}, or focus on the uncertainty from the observation noise, referred to as aleatoric uncertainty~\cite{KendallG17}. 
Following recent works~\cite{Gast018, YinDY19}, we focus on aleatoric uncertainty and how to train a generalizable uncertainty estimate in the context of self-supervised training. 
Wannenwetsch~\etal~\cite{ProbFlow} propose ProbFlow, a probabilistic approach applicable to energy-based optical flow algorithms~\cite{Barron94performanceof, RevaudWHS15, Sun2014}. 
Gast~\etal~\cite{Gast018} propose probabilistic output layers that require only minimal changes to existing networks. 
Yin~\etal~\cite{YinDY19} introduce HD$^3$F, a method to estimate uncertainty locally at multiple spatial scales and aggregate the results. 
While these approaches are carefully designed for optical flow data and restricted to small displacements, we consider the more general setting of estimating reliable confidence values for dense geometric matching, applicable to \eg pose-estimation and 3D reconstruction. This brings additional challenges, including coping with significant appearance changes and large geometric transformations.

\section{Our Approach: PDC-Net}
We introduce PDC-Net, a method for estimating the dense flow field relating two images, coupled with a robust pixel-wise confidence map. The later indicates the reliability and accuracy of the flow prediction, which is necessary for pose estimation, image manipulation, and 3D-reconstruction tasks.

\subsection{Probabilistic Flow Regression}
\label{subsec:proba-model}

We formulate dense correspondence estimation with a probabilistic model, which provides a framework for learning both the flow and its confidence in a unified formulation.
For a given image pair $X = \left(I^q, I^r \right)$ of spatial size $H \times W$, the aim of dense matching is to estimate a flow field $Y \in \mathbb{R}^{H \times W \times 2}$ relating the reference $I^r$ to the query $I^q$. Most learning-based methods address this problem by training a network $F$ with parameters $\theta$ that directly predicts the flow as $Y = F(X; \theta)$. However, this does not provide any information about the confidence of the prediction. 

Instead of generating a single flow prediction $Y$, our goal is to learn the conditional probability density $p(Y| X; \theta)$ of a flow $Y$ given the input $X$. This is generally achieved by letting a network predict the parameters $\Phi(X; \theta)$ of a family of distributions $p(Y | X; \theta) = p(Y|\Phi(X; \theta)) = \prod_{ij} p(y_{ij} | \varphi_{ij}(X; \theta))$. To ensure a tractable estimation, conditional independence of the predictions at different spatial locations $(i,j)$ is generally assumed. We use $y_{ij} \in \reals^2$ and $\varphi_{ij} \in \reals^n$ to denote the flow $Y$ and predicted parameters $\Phi$ respectively, at the spatial location $(i,j)$. In the following, we generally drop the sub-script $ij$ to avoid clutter. 

Compared to the direct approach $Y = F(X; \theta)$, the generated parameters $\Phi(X; \theta)$ of the predictive distribution can encode richer information about the flow prediction, including its uncertainty. In probabilistic regression techniques for optical flow~\cite{Gast018, IlgCGKMHB18} and a variety of other tasks~\cite{KendallG17, Shen2020, Walz2020}, this is most commonly performed by predicting the \emph{variance} of the estimate $y$.
In these cases, the predictive density $p(y|\varphi)$ is modeled using Gaussian or Laplace distributions. In the latter case, the density is given by,
\begin{equation}
\label{eq:laplace}
\mathcal{L}(y| \mu, \sigma^2) =\frac{1}{\sqrt{2 \sigma_u^2}} e^{-\sqrt{\frac{2}{\sigma_u^2}}|u-\mu_u|} . \frac{1}{\sqrt{2 \sigma_v^2}}
e^{-\sqrt{\frac{2}{\sigma_v^2}}|v-\mu_v|}
\end{equation}
where the components $u$ and $v$ of the flow vector $y=(u,v) \in \mathbb{R}^2$ are modelled with two conditionally independent Laplace distributions. The mean $\mu=[\mu_u, \mu_v]^T \in \mathbb{R}^2$ and variance $\sigma^2=[\sigma^2_u, \sigma^2_v]^T \in \mathbb{R}^2$ of the distribution $p(y|\varphi) = \mathcal{L}(y | \mu, \sigma^2)$ are predicted by the network as $(\mu, \sigma^2) = \varphi(X; \theta)$ at every spatial location. 

\begin{figure}[t]
\centering%

\includegraphics[width=0.40\textwidth]{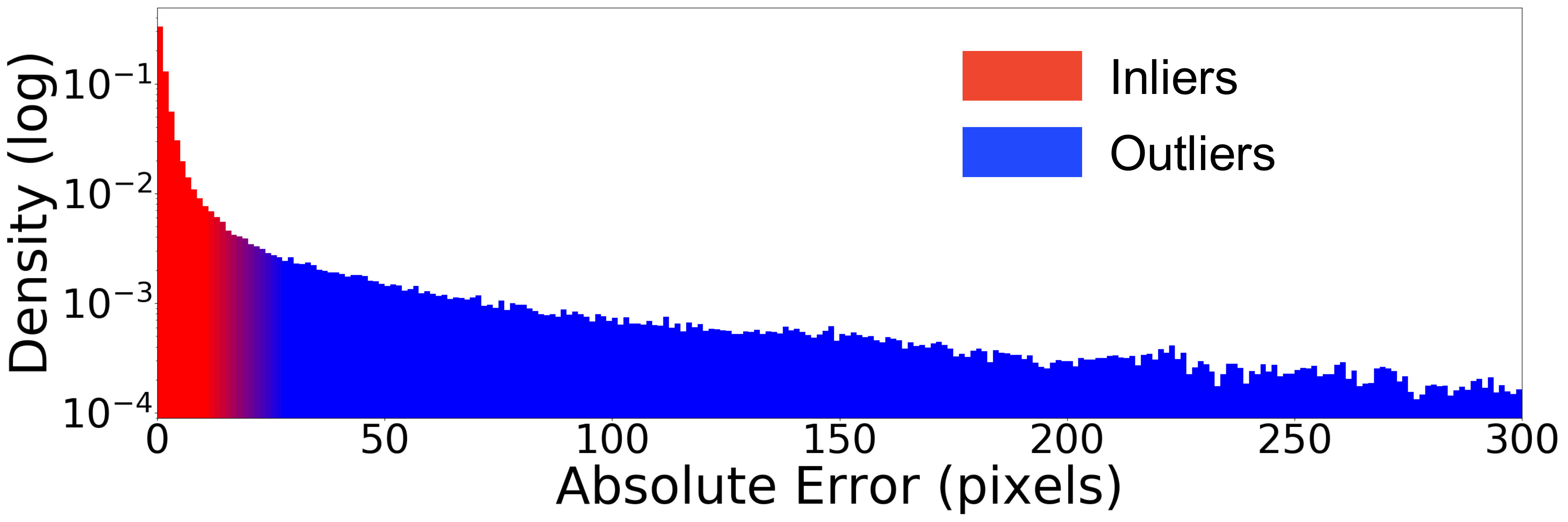}
\caption{Distribution of errors 
$|\widehat{y}-y|$ on MegaDepth~\cite{megadepth} between the flow $\widehat{y}$ estimated by GLU-Net~\cite{GLUNet} and the ground-truth $y$. 
}
\vspace{-4mm}
\label{fig:distribution}
\end{figure}

\begin{figure*}[t]
\centering%
\newcommand{\wid}{\textwidth}
\includegraphics[width=\textwidth]{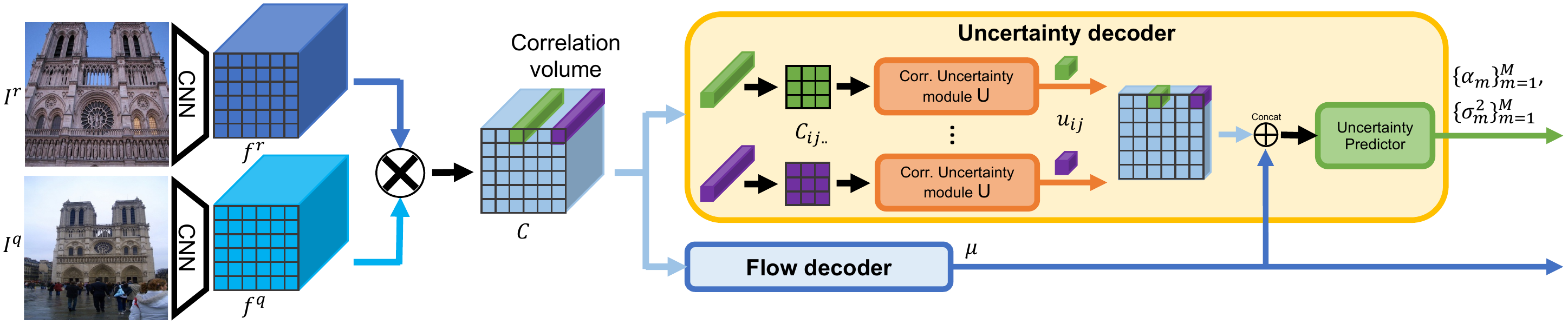}
\vspace{-5mm}\caption{The proposed architecture for flow and uncertainty estimation. The correlation uncertainty module $U_\theta$ independently processes each 2D-slice $C_{ij\cdot\cdot}$ of the correlation volume. Its output is combined with the estimated mean flow $\mu$ to predict the weight $\{\alpha_m\}_1^M$ and variance $\{\sigma^2_m\}_1^M$ parameters of our constrained mixture model \eqref{eq:mixture}-\eqref{eq:scalepred}.}\vspace{-4mm} 
\label{fig:arch}
\end{figure*}

\subsection{Constrained Mixture Model Prediction}
\label{sec:constained-mixture}

Fundamentally, the goal of probabilistic deep learning is to achieve a predictive model $p(y|X;\theta)$ that coincides with empirical probabilities as well as possible. We can get important insights into this problem by studying the empirical error distribution of a state-of-the-art matching model, in this case GLU-Net~\cite{GLUNet}, as shown in Fig.~\ref{fig:distribution}. 
Errors can be categorized into two populations: inliers (in red) and outliers (in blue). Current probabilistic methods~\cite{Gast018, IlgCGKMHB18,abs-2010-04367} mostly rely on a Laplacian model \eqref{eq:laplace} of $p(y|X;\theta)$. Such a model is effective for correspondences which are easily estimated to be either inliers or outliers with \emph{high certainty}, by predicting a low or high variance respectively. 
However, often the network is not certain whether a match is an inlier or outlier. A single Laplace can only predict an intermediate variance, which does not faithfully represent the more complicated uncertainty pattern in this case.

\parsection{Mixture model}
To achieve a flexible model capable of fitting more complex distributions, we parametrize $p(y | X; \theta)$ with a mixture model. In general, we consider a distribution consisting of $M$ components, 
\begin{equation}
\label{eq:mixture}
p\left(y | \varphi \right)=\sum_{m=1}^{M} \alpha_{m}  \mathcal{L}\left(y |\mu, \sigma^2_m\right) \,.
\end{equation}
While we have here chosen Laplacian components \eqref{eq:laplace}, any simple density function can be used. The scalars $\alpha_{m} \geq 0$ control the weight of each component, satisfying $\sum_{m=1}^{M} \alpha_{m} = 1$. Note that all components have the same mean $\mu$, which can thus be interpreted as the estimated flow vector, but different variances $\sigma^2_m$. 
The distribution \eqref{eq:mixture} is therefore unimodal, but can capture more complex uncertainty patterns. In particular, it allows to predict the probability of inlier (red) and outlier (blue) matches (Fig.~\ref{fig:distribution}), each modeled by separate Laplace components.

\parsection{Mixture constraints}
In general, we now consider a network $\Phi$ that, for each pixel location, predicts the mean flow $\mu$ along with the variance $\sigma^2_m$ and weight $\alpha_m$ of each component, as $\big( \mu, (\alpha_m )_{m=1}^M, ( \sigma^2_m  )_{m=1}^M  \big) = \varphi(X; \theta)$. 
However, a potential issue when predicting the parameters of a mixture model is its permutation invariance. That is, the predicted distribution \eqref{eq:mixture} is unchanged even if we change the order of the individual components. This can cause confusion in the learning, since the network first needs to \emph{decide} what each component should model before estimating the individual weights $\alpha_m$ and variances $\sigma^2_m$. 

We propose a model that breaks the permutation invariance of the mixture \eqref{eq:mixture}, which simplifies the learning and greatly improves the robustness of the estimated uncertainties. In essence, each component $m$ is tasked with modeling a specified range of variances $\sigma^2_m$. We achieve this by constraining the mixture \eqref{eq:mixture} as,
\begin{equation}
\label{eq:constraint}
0 < \beta_1^- \leq \sigma^2_1 \leq \beta_1^+ \leq \beta_2^- \leq \sigma^2_2 \leq \ldots \leq \sigma^2_M \leq \beta_M^+ 
\end{equation}
For simplicity, we here assume a single variance parameter $\sigma^2_m$ for both the $u$ and $v$ directions in \eqref{eq:laplace}.
The constants $\beta_m^-, \beta_m^+$ specify the range of variances $\sigma^2_m$. Intuitively, each component is thus responsible for a different range of uncertainties, roughly corresponding to different regions in the error distribution in Fig.~\ref{fig:distribution}. In particular, component $m=1$ accounts for the most accurate predictions, while component $m=M$ models the largest errors and outliers. 
To enforce the constraint \eqref{eq:constraint}, we first predict an unconstrained value $h_m \in \reals$, which is then mapped to the given range as,
\begin{equation}
\label{eq:scalepred}
\sigma^2_m = \beta_{m}^- + (\beta_m^+ - \beta_{m}^-)\, \text{Sigmoid}(h_m) .
\end{equation}
The constraint values $\beta_m^+, \beta_m^-$ can either be treated as hyper-parameters or learned end-to-end alongside $\theta$. 

\begin{figure}[b]
\vspace{-3mm}
\centering%
{\includegraphics[width=0.59\columnwidth]{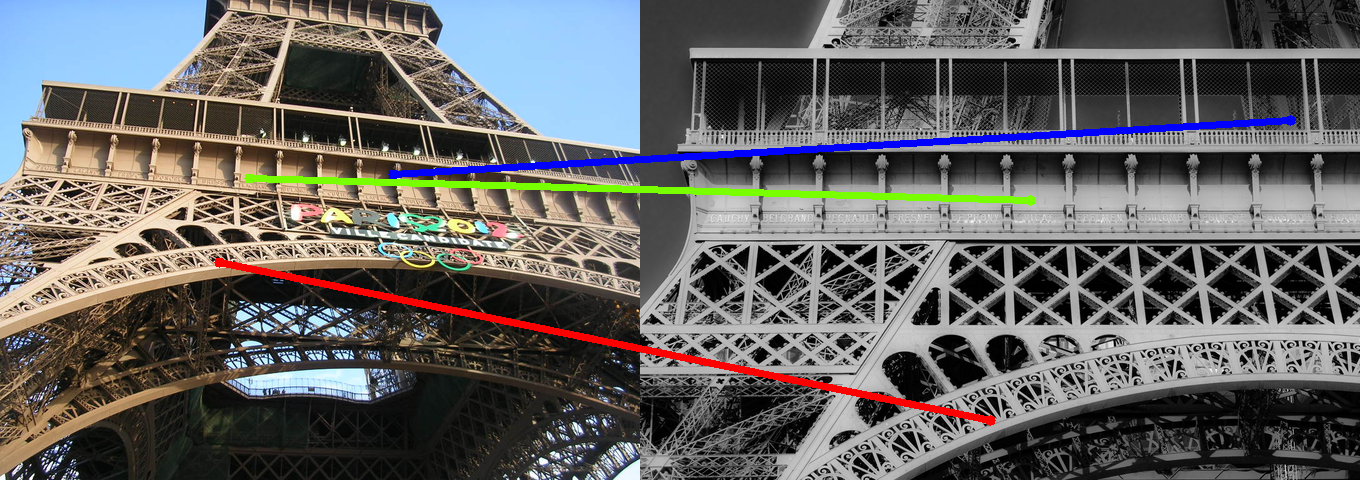}}~%
{\includegraphics[width=0.31\columnwidth]{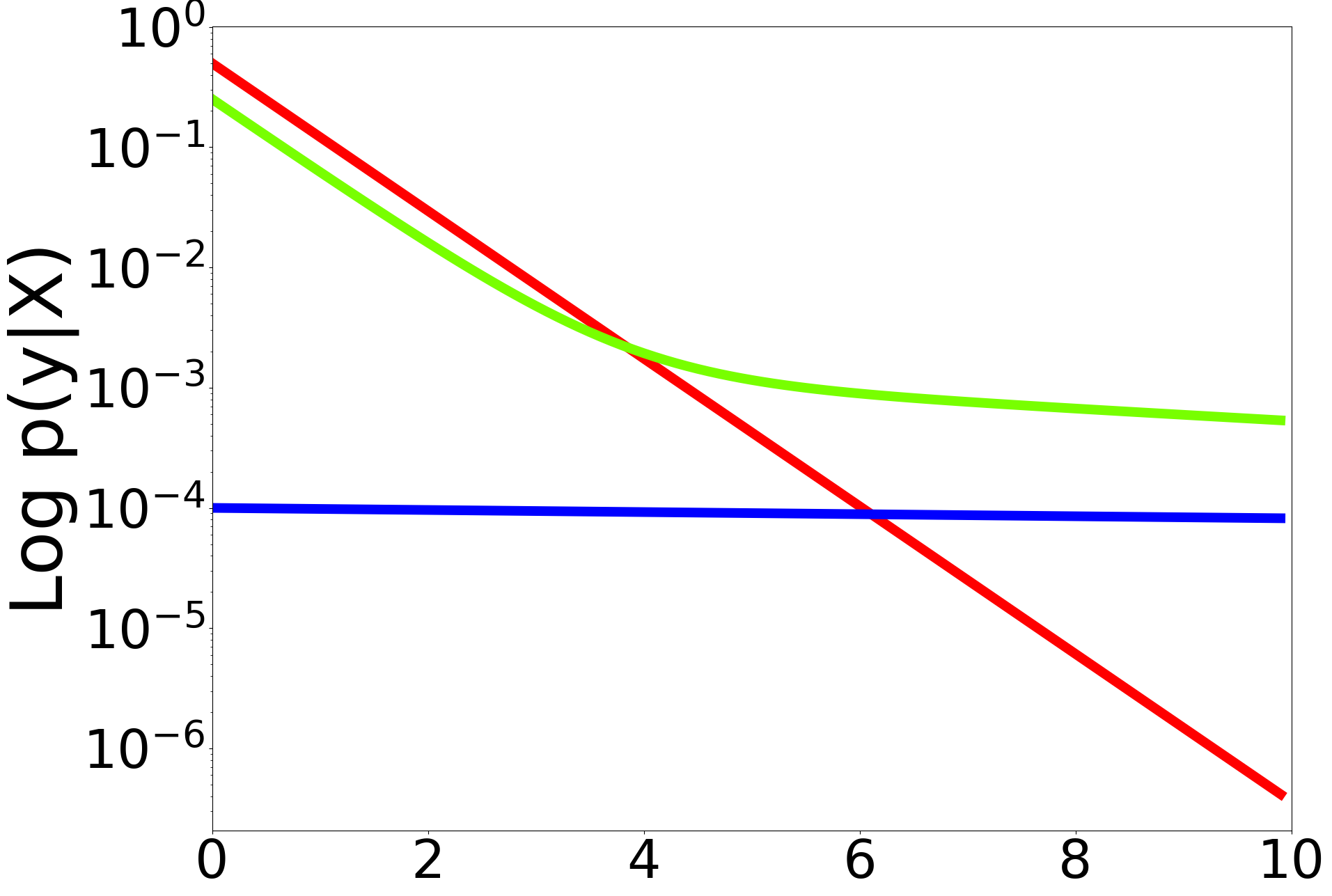}}
\caption{Predictive log-distr.\ $\log p(y|X)$ \eqref{eq:mixture}-\eqref{eq:constraint} for an 
inlier (red), outlier (blue), and ambiguous (green) match. 
Our mixture model faithfully represents the uncertainty also in the latter case.
}
\label{fig:predictive}
\end{figure}

Lastly, we emphasize an interesting interpretation of our constrained mixture formulation \eqref{eq:mixture}-\eqref{eq:constraint}. Note that the predicted weights $\alpha_m$, in practice obtained through a final SoftMax layer, represent the probabilities of each component $m$. Our network therefore effectively \emph{classifies} the flow prediction at each pixel into the separate uncertainty intervals \eqref{eq:constraint}. 
We visualize in Fig.~\ref{fig:predictive} the predictive log-distribution with $M=2$ for three cases. The red and blue matches are with certainty predicted as inlier and outlier respectively, thus requiring only a single active component. In ambiguous cases (green), our mixture model \eqref{eq:mixture}-\eqref{eq:constraint} predicts the probability of inlier vs.\ outlier, giving a better fit compared to a single-component alternative.
As detailed next, our network learns this ability without any extra supervision.

\parsection{Training objective}
As customary in probabilistic regression~\cite{prdimp, Gast018, ebmregECCV2020, IlgCGKMHB18, KendallG17, Shen2020, VarameshT20, Walz2020}, we train our method using the negative log-likelihood as the only objective. For one input image pair $X = \left(I^q, I^r \right)$ and corresponding ground-truth flow $Y$, the objective is given by
\begin{equation}
\label{eq:nll}
    - \log p\big(Y|\Phi(X;\theta)\big) = - \sum_{ij} \log p\big(y_{ij} | \varphi_{ij}(X;\theta)\big) \,.
\end{equation}
In Appendix~\ref{sec-sup:training-loss}, we provide efficient analytic expressions of the loss \eqref{eq:nll} for our constrained mixture \eqref{eq:mixture}-\eqref{eq:scalepred}, that also ensure numerical stability. 
As detailed in Sec.~\ref{subsec:arch}, we can train our final model using either a self-supervised strategy where $X$ and $Y$ are generated by artificial warping, using real sparse ground-truth $Y$, or a combination of both. Next, we present the architecture of our network $\Phi$ that predicts the parameters of our constrained mixture model \eqref{eq:mixture}.

\begin{figure*}[t]
\centering%
\newcommand{\wid}{0.16\textwidth}
\vspace{-3mm}
\subfloat[Query image \label{fig:arch-visual-query}]{\includegraphics[width=\wid]{  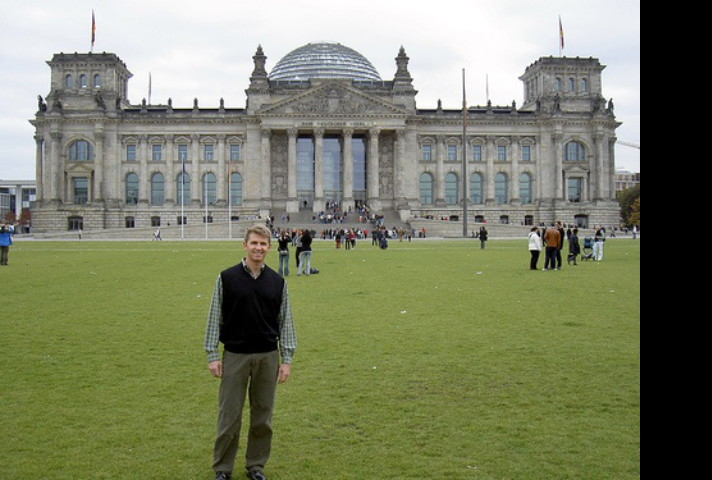}}~%
\subfloat[Reference image \label{fig:arch-visual-ref}]{\includegraphics[width=\wid]{  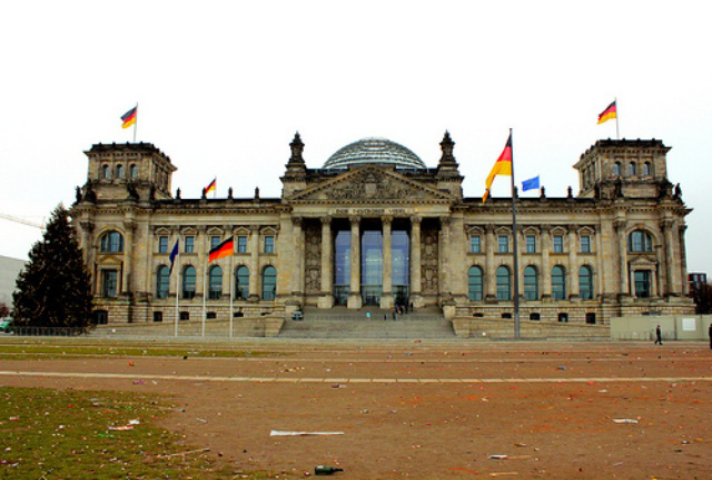}}~%
\subfloat[\centering Common decoder\label{fig:arch-visual-common}]{\includegraphics[width=\wid]{  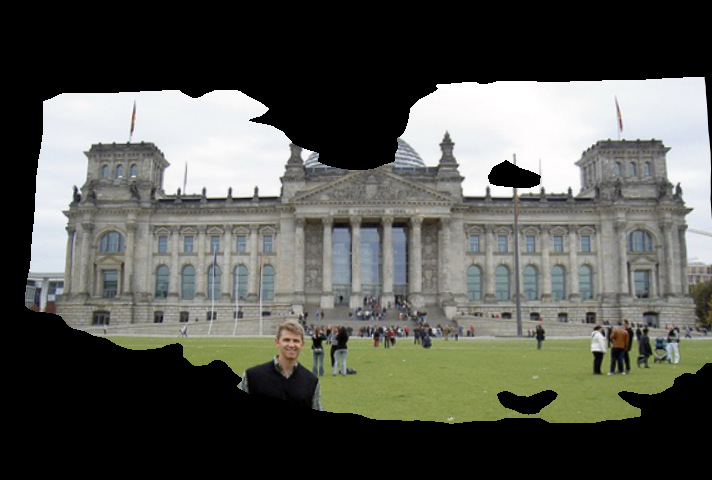}}~%
\subfloat[\centering Our decoder \label{fig:arch-visual-sep}]{\includegraphics[width=\wid]{  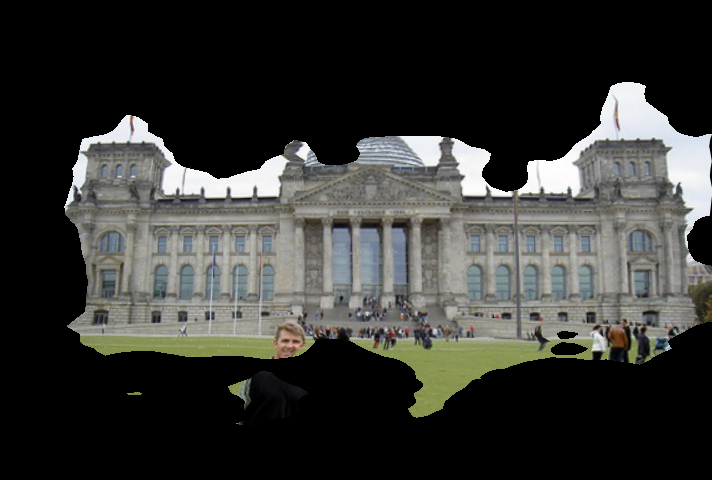}}~%
\subfloat[\centering Our decoder and data \label{fig:arch-visual-sep-pertur}]{\includegraphics[width=\wid]{  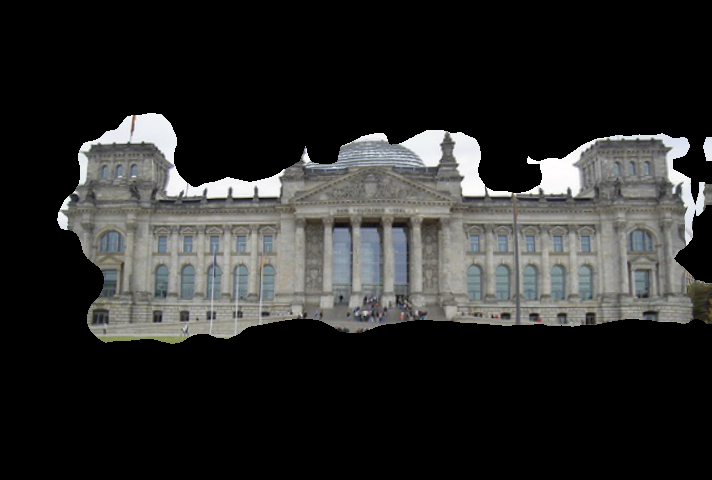}}~%
\subfloat[\centering RANSAC-Flow \label{fig:arch-visual-ransac}]{\includegraphics[width=\wid]{  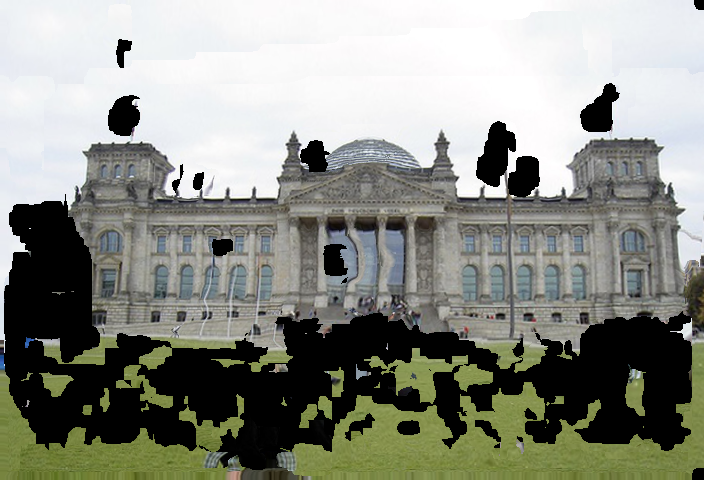}} 
\vspace{-3mm}\caption{Visualization of the estimated uncertainties by masking the warped query image to only show the confident flow predictions. The standard approach \protect\subref{fig:arch-visual-common} uses a common decoder for both flow and uncertainty estimation. It generates overly confident predictions in the sky and grass. The uncertainty estimates are substantially improved in \protect\subref{fig:arch-visual-sep}, when using the proposed architecture described in Sec.~\ref{sec:uncertainty-arch}. 
Adding the flow perturbations for self-supervised training (Sec.~\ref{subsec:perturbed-data}) further improves the robustness and generalization of the uncertainties \protect\subref{fig:arch-visual-sep-pertur}. For reference, we also visualize the flow and confidence mask \protect\subref{fig:arch-visual-ransac} predicted by the recent state-of-the-art approach RANSAC-Flow~\cite{RANSAC-flow}.
}\vspace{-4mm}
\label{fig:arch-visual}
\end{figure*}

\subsection{Uncertainty Prediction Architecture}
\label{sec:uncertainty-arch}

Our aim is to predict an uncertainty value that quantifies the \emph{reliability} of a proposed correspondence or flow vector. Crucially, the uncertainty prediction needs to \emph{generalize} well to real scenarios, not seen during training. 
However, this is particularly challenging in the context of self-supervised training, which relies on synthetically warped images or animated data. Specifically, when trained on simple synthetic motion patterns, such as homography transformations, the network learns to heavily rely on global smoothness assumptions, which do not generalize well to more complex settings. As a result, the network learns to \emph{confidently} interpolate and extrapolate the flow field to regions where no robust match can be found. 
Due to the significant distribution shift between training and test data, the network thus also infers confident, yet highly erroneous predictions in homogeneous regions on real data. 
In this section, we address this problem by carefully designing an architecture that greatly limits the risk of the aforementioned issues. Our architecture is visualized in Figure \ref{fig:arch}.

Current state-of-the-art dense matching architectures rely on feature correlation layers. Features $f$ are extracted at resolution $h \times w$ from a pair of input images, and densely correlated either globally or within a local neighborhood of size $d$. In the later case, the output correlation volume is best thought of as a 4D tensor $C \in \reals^{h \times w \times d \times d}$. Computed as dense scalar products $C_{ijkl} = (f_{ij}^r)\tp f_{i+k,j+l}^q$, it encodes the deep feature similarity between a location $(i,j)$ in the reference frame $I^r$ and a displaced location $(i+k,j+l)$ in the query $I^q$. Standard flow architectures process the correlation volume by first vectorizing the last two dimensions, before applying a sequence of convolutional layers over the \emph{reference coordinates $(i,j)$} in order to predict the final flow.  

\parsection{Correlation uncertainty module}
The straightforward strategy for implementing the parameter predictor $\Phi(X;\theta)$ is to simply  increase the number of output channels to include all parameters of the predictive distribution. 
However, this allows the network to rely primarily on the local neighborhood when estimating the flow and confidence at location $(i,j)$, and thus to ignore the actual reliability of the match and appearance information at the specific location. It results in over-smoothed and overly confident predictions, unable to identify ambiguous and unreliable matching regions, such as the sky. This is visualized in Fig.~\ref{fig:arch-visual-common}.

We instead design an architecture that assesses the uncertainty at a specific location $(i,j)$, without relying on neighborhood information. We note that the 2D slice $C_{ij\cdot\cdot} \in \reals^{d\times d}$ encapsulates rich information about the matching ability of location $(i,j)$, in the form of a confidence map. 
In particular, it encodes the distinctness, uniqueness, and existence of the correspondence.
We therefore create a \emph{correlation uncertainty decoder} $U_{\theta}$ that independently reasons about each correlation slice as $U_\theta(C_{ij\cdot\cdot})$. In contrast to standard decoders, the convolutions are therefore applied over \emph{the displacement dimensions $(k,l)$}. Efficient parallel implementation is ensured by moving the first two dimensions of $C$ to the batch dimension using a simple tensor reshape. Our strided convolutional layers then gradually decrease the size $d \times d$ of the displacement dimensions $(k,l)$ until a single vector $u_{ij} = U_\theta(C_{ij\cdot\cdot}) \in \reals^n$ is achieved for each spatial coordinate $(i,j)$ (see Fig.~\ref{fig:arch}). 

\parsection{Uncertainty predictor} 
The cost volume does not capture uncertainty arising at motion boundaries, crucial for real data with independently moving objects.
We thus additionally integrate predicted flow information in the estimation of its uncertainty. In practise, we concatenate the estimated mean flow $\mu$ with the output of the correlation uncertainty module $U_{\theta}$, and process it with multiple convolution layers. It outputs all parameters of the mixture \eqref{eq:mixture}, except for the mean flow $\mu$ (see Fig.~\ref{fig:arch}). 
As shown in Fig.~\ref{fig:arch-visual-sep}, our uncertainty decoder, comprised of the correlation uncertainty module and the uncertainty predictor, successfully masks out most of the inaccurate and unreliable matching regions.

\subsection{Data for Self-supervised Uncertainty}
\label{subsec:perturbed-data}

While designing a suitable architecture greatly alleviates the uncertainty generalization issue, the network still tends to rely on global smoothness assumptions and interpolation, especially around object boundaries (see Fig.~\ref{fig:arch-visual-sep}). While this learned strategy indeed minimizes the Negative Log Likelihood loss \eqref{eq:nll} on self-supervised training samples, it does not generalize to real image pairs. 
In this section, we further tackle this problem from the data perspective in the context of self-supervised learning. 

We aim at generating less predictable synthetic motion patterns than simple homography transformations, to prevent the network from primarily relying on interpolation. This forces the network to focus on the appearance of the image region in order to predict its motion and uncertainty.
Given a base flow $\tilde{Y}$ relating $\tilde{I}^r$ to $\tilde{I}^q$ and representing a simple transformation such as a homography as in prior works~\cite{Melekhov2019, Rocco2017a, GOCor, GLUNet}, we create a residual flow $\epsilon = \sum_i \varepsilon_i$, by adding small local perturbations $\varepsilon_i$. The query image $I^q = \tilde{I}^q$ is left unchanged  while the reference $I^r$ is generated by warping $\tilde{I}^r$ according to the residual flow $\epsilon$. 
The final perturbed flow map $Y$ between $I^r$ and $I^q$ is achieved by composing the base flow $\tilde{Y}$ with the residual flow $\epsilon$. 

An important benefit of introducing the perturbations $\varepsilon$ is to teach the network to be uncertain in regions where it cannot identify them. Specifically, in homogeneous regions such as the sky, the perturbations do not change the appearance of the reference ($I^r \approx \tilde{I}^r$) and are therefore unnoticed by the network. However, since the perturbations break the global smoothness of the synthetic flow, the flow errors on those pixels will be higher. In order to decrease the loss \eqref{eq:nll}, the network will thus need to estimate a larger uncertainty for these regions. 
We show the impact of introducing the flow perturbations for self-supervised learning in Fig.~\ref{fig:arch-visual-sep-pertur}.

\subsection{Geometric Matching Inference}
\label{sec:geometric-inference}

In real settings with extreme view-point changes, flow estimation is prone to failing.
Our confidence estimate can be used to improve the robustness of matching networks to such cases. Particularly, our approach offers the opportunity to perform multi-stage flow estimation on challenging image pairs, without any additional network components.

\parsection{Confidence value} From the predictive distribution $p(y|\varphi(X;\theta))$, we aim at extracting a single confidence value, encoding the reliability of the corresponding predicted flow vector $\mu$. 
Previous probabilistic regression methods mostly rely on the variance as a confidence measure~\cite{Gast018, IlgCGKMHB18,  KendallG17, Walz2020}. However, we observe that the variance can be sensitive to outliers. Instead, we compute the probability $P_R$ of the true flow being within a radius $R$ of the estimated mean flow $\mu$. This is expressed as,
\begin{equation}
    P_R = P(|y - \mu| < R) = \int_{\{y\in \reals^2: |y - \mu| < R\}}\! p(y|\varphi) dy.
\end{equation}
Compared to the variance, the probability value $P_R$ also provides a more interpretable measure of the uncertainty. 

\parsection{Multi-stage refinement strategy}  
For extreme view-point changes with large scale or perspective variations, it is particularly difficult to infer the correct motion field in a single network pass. While this is partially alleviated by multi-scale architectures, it remains a major challenge in geometric matching. 
Our approach allows to split the flow estimation process into two parts, the first estimating a simple transformation, which is then used as initialization to infer the final, more complex transformation. 

One of the major benefits of our confidence estimation is the ability to \emph{identify} the set of accurate matches from the densely estimated flow field. After a first forward network pass, these accurate correspondences can be used to estimate a coarse transformation relating the image pair, such as a homography transformation. A second forward pass can then be applied to the coarsely aligned image pair, and the final flow field is constructed as a composition of the fine flow and the homography transform.
While previous works also use multi-stage refinement~\cite{Rocco2017a, RANSAC-flow}, our approach is much simpler, applying the \emph{same} network in both stages and benefiting from the internal confidence estimation.
\section{Experimental results}
\label{sec:exp}

We integrate our approach into a generic pyramidal correspondence network and perform comprehensive experiments on multiple geometric matching and optical flow datasets. We also show that our method can be used for various tasks, including pose estimation and dense 3D reconstruction.  
Further results, analysis, visualizations and implementation details are provided in the Appendix.

\begin{figure}[b]
\centering%
\vspace{-3mm}\newcommand{\wid}{0.15\textwidth}
\includegraphics[width=0.35\textwidth]{   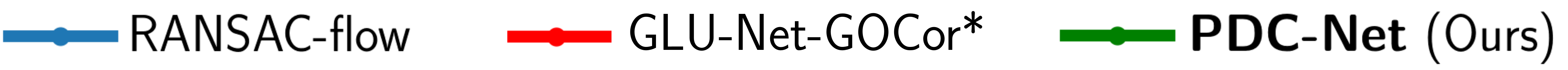} \\
\vspace{-3mm}\subfloat{\includegraphics[width=\wid]{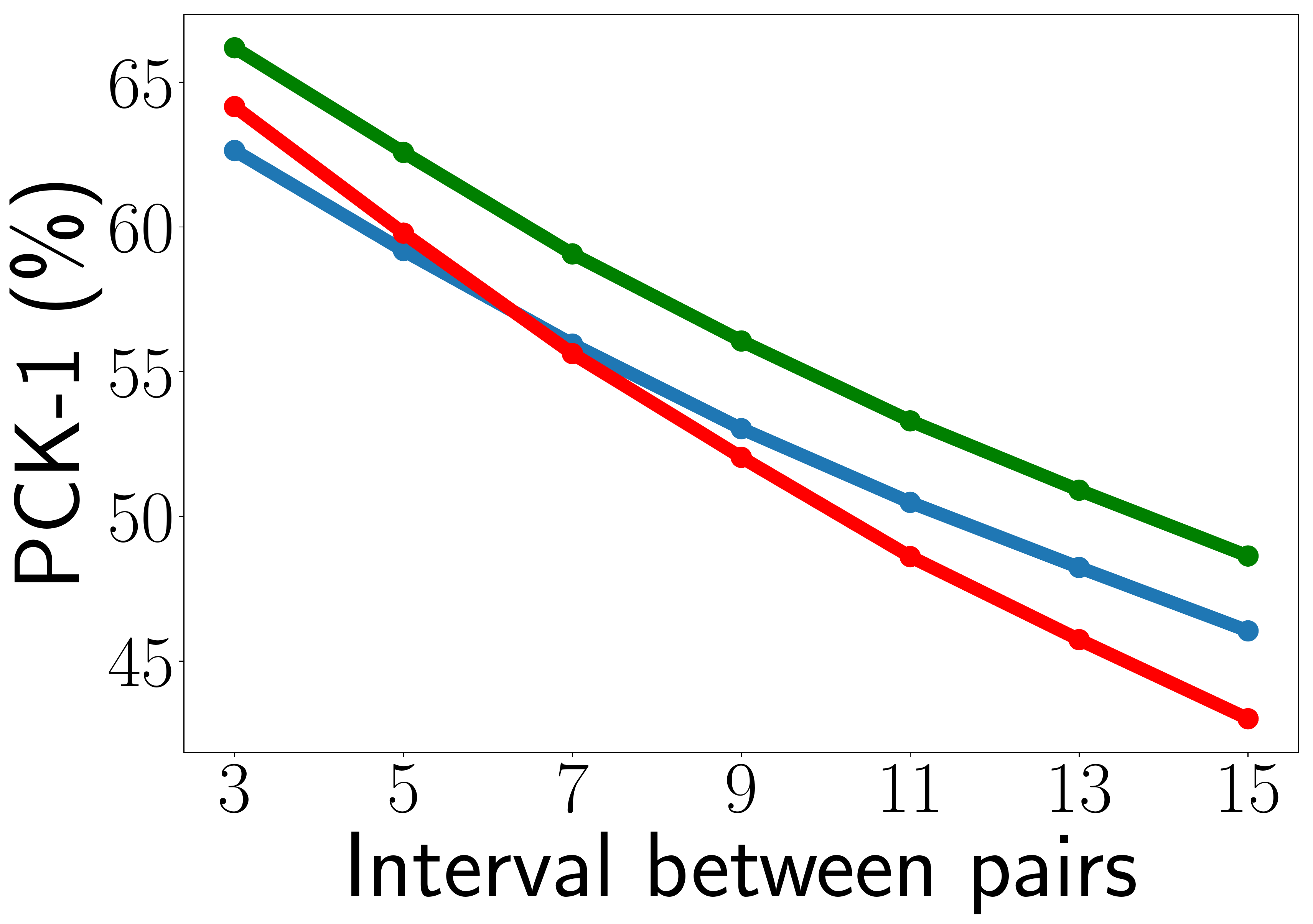}}~%
\subfloat{\includegraphics[width=\wid]{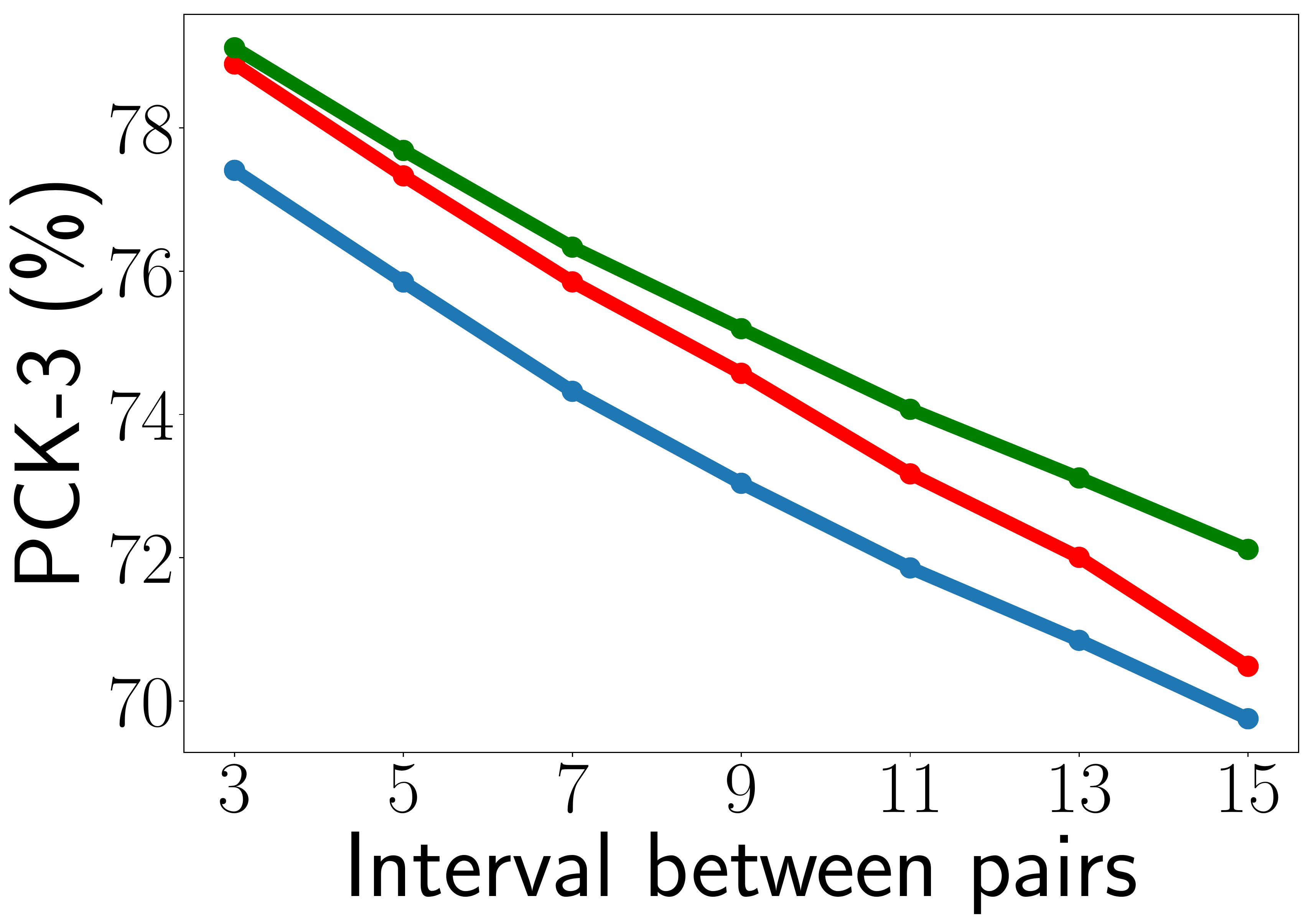}}~%
\subfloat{\includegraphics[width=\wid]{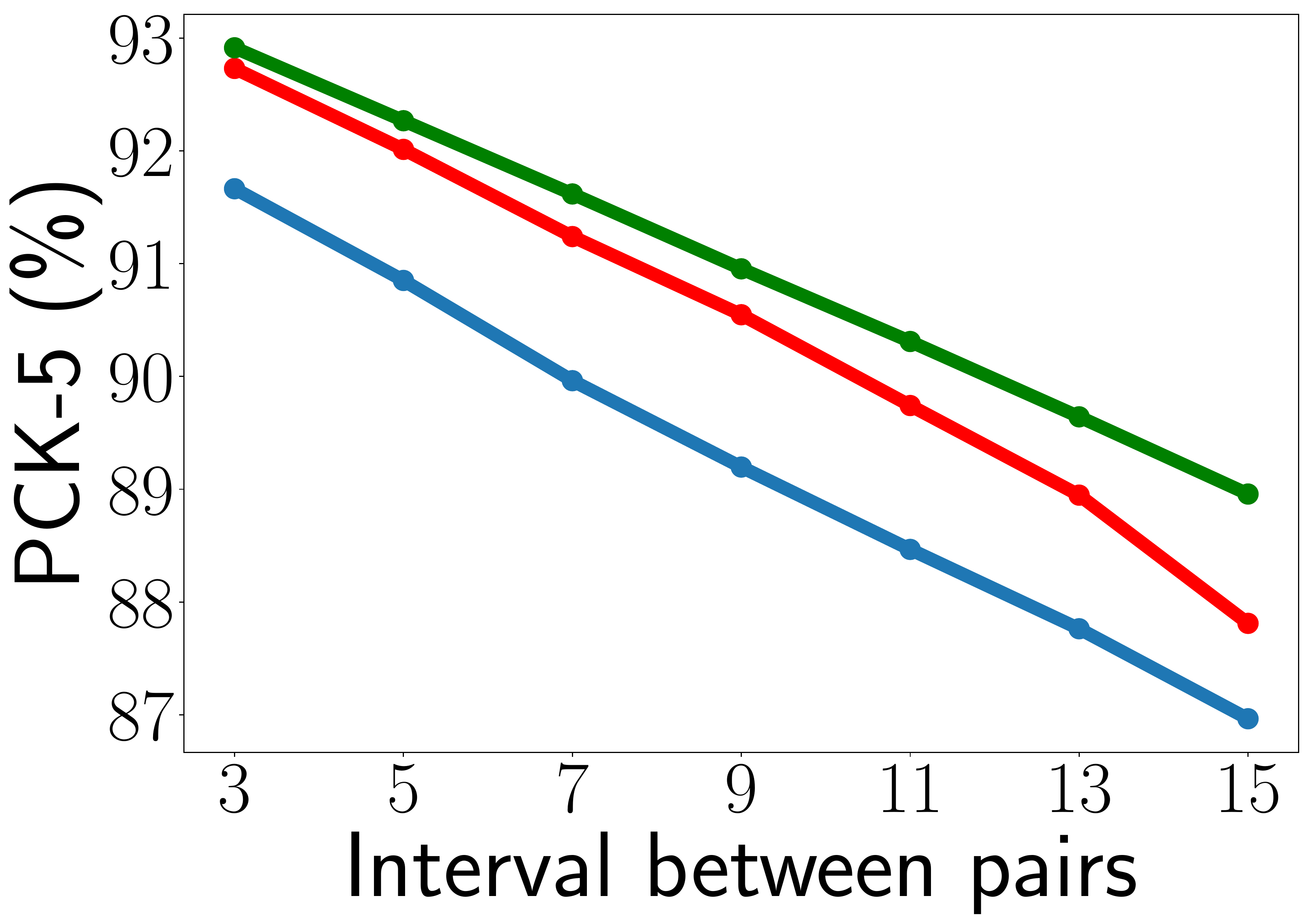}}~%
\caption{Results on ETH3D~\cite{ETH3d}.  PCK-1 (left), PCK-3 (center) and PCK-5 (right) are plotted w.r.t. the inter-frame interval length.}
\label{fig:ETH3D}
\end{figure}

\subsection{Implementation Details}
\label{subsec:arch}

We adopt the recent GLU-Net-GOCor~\cite{GOCor, GLUNet} as our base architecture. It consists in a four-level pyramidal network operating at two image resolutions and employing a VGG-16 network~\cite{Chatfield14} pre-trained on ImageNet for feature extraction. At each level, we add our uncertainty decoder (Sec.~\ref{sec:uncertainty-arch}) and propagate the uncertainty prediction to the next level. 
We model the probability distribution $p\left(y | \varphi \right)$ with a constrained mixture (Sec.~\ref{sec:constained-mixture}) with $M=2$ Laplace components, where the first is fixed to $\sigma^2_1 = \beta_1^- = \beta_1^+ = 1$ to represent the very accurate predictions, while the second models larger errors and outliers, as $ 2 = \beta_2^- \leq \sigma^2_2 \leq \beta_2^+$, where $\beta_2^+$ is set to the square of the training image size.

Our training consists of two stages.
First, we follow the self-supervised training procedure of~\cite{GOCor, GLUNet}. Random homography transformations are applied to images compiled from different sources to ensure diversity. For better compatibility with real 3D scenes and moving objects, the data is further augmented with random independently moving objects from the COCO~\cite{coco} dataset. 
We further apply our perturbation strategy described in Sec.~\ref{subsec:perturbed-data}.
In the second stage, we extend the self-supervised data with real image pairs with sparse ground-truth correspondences from the MegaDepth dataset~\cite{megadepth}. We additionally fine-tune the backbone feature extractor.
For fair comparison, we also train a version of GLU-Net-GOCor, denoted GLU-Net-GOCor*, using the same settings and data. 

For datasets with very extreme geometric transformations, we also report using a multi-scale strategy. In particular, we extend our two-stage refinement approach (Sec.~\ref{sec:geometric-inference}) by resizing the reference image to different resolutions. The resulting image pairs are passed through the network and we fit a homography for each pair, using our predicted flow and uncertainty map. We select the homography with the highest percentage of inliers, and scale it to the images original resolutions. The original image pair is then coarsely aligned and from there we follow the same procedure, as explained in Sec.~\ref{sec:geometric-inference}. We refer to this option as Multi Scale (MS).

\begin{table}[t]
\centering
\resizebox{\columnwidth}{!}{%
\begin{tabular}{lccc|ccc} \toprule
& \multicolumn{3}{c}{\textbf{MegaDepth}} & \multicolumn{3}{c}{\textbf{RobotCar}} \\
& PCK-1  & PCK-3  & PCK-5 & PCK-1  & PCK-3 & PCK-5 \\ \midrule
SIFT-Flow~\cite{LiuYT11} & 8.70 & 12.19 & 13.30 & 1.12 & 8.13 & 16.45\\
NCNet~\cite{Rocco2018b} & 1.98 & 14.47 & 32.80 & 0.81 & 7.13 & 16.93 \\
DGC-Net~\cite{Melekhov2019} & 3.55 & 20.33 & 32.28 & 1.19 & 9.35 & 20.17 \\
GLU-Net~\cite{GLUNet} & 21.58 & 52.18 & 61.78 & 2.30 & 17.15 & 33.87 \\
GLU-Net-GOCor~\cite{GOCor} &  37.28 &  61.18  & 68.08 & 2.31 & 17.62 & 35.18 \\ 
\midrule
RANSAC-Flow (MS)~\cite{RANSAC-flow} & 53.47 & 83.45 & 86.81 & 2.10 & 16.07 & 31.66 \\
GLU-Net-GOCor* & 57.86 & 78.62 & 82.30 & 2.33 & 17.21 & 33.67 \\
\textbf{PDC-Net} & 70.75 & 86.51 & 88.00 & 2.54 & \textbf{18.97} & \textbf{36.37}\\
\textbf{PDC-Net} (MS) & \textbf{71.81} & \textbf{89.36} & \textbf{91.18} & \textbf{2.58} & 18.87 & 36.19\\
\bottomrule
\end{tabular}%
}\vspace{1mm}\caption{PCK (\%) results on sparse correspondences of the MegaDepth~\cite{megadepth} and RobotCar~\cite{RobotCar, RobotCarDatasetIJRR} datasets.
}\vspace{-5.5mm}
\label{tab:megadepth}
\end{table}

\subsection{Geometric Correspondences and Flow}
\label{subsec:correspondence-est}

We first evaluate our PDC-Net in terms of the quality of the predicted flow field. 

\parsection{Datasets and metrics} We evaluate on standard datasets with sparse ground-truth, namely the \textbf{RobotCar}~\cite{RobotCar, RobotCarDatasetIJRR}, \textbf{MegaDepth}~\cite{megadepth} and \textbf{ETH3D}~\cite{ETH3d} datasets. RobotCar depicts outdoor road scenes, taken under different weather and lighting conditions. 
Images are particularly challenging due to their numerous textureless regions. 
MegaDepth images show extreme view-point and appearance variations. Finally, ETH3D represents indoor and outdoor scenes captured from a moving hand-held camera.
For RobotCar and MegaDepth, we evaluate on the correspondences provided by~\cite{RANSAC-flow}, which includes approximately 340M and 367K ground-truth matches respectively. For ETH3D, we follow the protocol of~\cite{GLUNet}, sampling image pairs at different intervals to analyze varying magnitude of geometric transformations, resulting in 600K to 1100K matches per interval. In line with~\cite{RANSAC-flow}, we employ the Percentage of Correct Keypoints at a given pixel threshold $T$ (PCK-$T$) as metric.

\parsection{Results}
In Tab.~\ref{tab:megadepth} we report results on MegaDepth and RobotCar.  Our method PDC-Net outperforms all previous works by a large margin at all PCK thresholds.
In particular, our approach is significantly more accurate and robust than the very recent RANSAC-Flow, which utilizes an extensive multi-scale (MS) search.
Interestingly, our uncertainty-aware probabilistic approach also outperforms the baseline GLU-Net-GOCor* in pure flow accuracy. 
This clearly demonstrates the advantages of casting the flow estimation as a probabilistic regression problem, advantages which are not limited to uncertainty estimation. It also substantially benefits the accuracy of the flow itself through a more flexible loss formulation.
In Fig.~\ref{fig:ETH3D}, we plot the PCKs on ETH3D. Our approach is consistently better than RANSAC-Flow and GLU-Net-GOCor* for all intervals.

\parsection{Generalization to optical flow} 
We additionally show that our approach generalizes well to accurate estimation of optical flow, even though it is trained for the very different task of geometric matching.
We use the established \textbf{KITTI} dataset~\cite{Geiger2013}, and evaluate according to the standard metrics, namely AEPE and F1. 
Since we do not fine-tune on KITTI, we show results on the training splits in Tab.~\ref{tab:optical-flow}.
Our approach outperforms all previous generic matching methods (upper part) by a large margin in terms of both F1 and AEPE. Surprisingly, PDC-Net also obtains better results than all optical flow methods (bottom part), even outperforming the recent RAFT~\cite{RAFT} on KITTI-2015.   

\begin{table}[t]
\centering
\resizebox{0.35\textwidth}{!}{%
\begin{tabular}{lcc|cc}
\toprule
             & \multicolumn{2}{c}{\textbf{KITTI-2012}} & \multicolumn{2}{c}{\textbf{KITTI-2015}} \\ 
  & AEPE  $\downarrow$            & F1   (\%)   $\downarrow$      & AEPE  $\downarrow$              & F1  (\%)  $\downarrow$ \\ \midrule
DGC-Net~\cite{Melekhov2019}  &  8.50  &   32.28 &  14.97   &      50.98 \\
GLU-Net~\cite{GLUNet} & 3.14 & 19.76 & 7.49 & 33.83 \\
GLU-Net-GOCor~\cite{GOCor} & 2.68 & 15.43 & 6.68 & 27.57 \\ 
RANSAC-Flow~\cite{RANSAC-flow} & - & - &  12.48 & - \\
GLU-Net-GOCor* & 2.26 & 10.23 & 5.58 & 18.76 \\
\textbf{PDC-Net} & \textbf{2.08} & \textbf{7.98} & \textbf{5.22} & \textbf{15.13} \\
\midrule 
PWC-Net~\cite{Sun2018} & 4.14 & 21.38 & 10.35 & 33.7  \\
LiteFlowNet~\cite{Hui2018} & 4.00 & - &  10.39 & 28.5  \\
HD$^3$F~\cite{YinDY19} & 4.65 & - & 13.17 & 24.0 \\
LiteFlowNet2~\cite{Hui2019} & 3.42 & - & 8.97 & 25.9 \\
VCN~\cite{VCN} & - & - & 8.36 & 25.1 \\
RAFT~\cite{RAFT} & - & - &  5.54 & 19.8  \\
\bottomrule
\end{tabular}%
}\vspace{1mm}\caption{Optical flow results on the training splits of KITTI~\cite{Geiger2013}. The upper part contains generic matching networks, while the bottom part lists specialized optical flow methods, not trained on kitti. }\vspace{-5mm}
\label{tab:optical-flow}
\end{table}

\begin{figure}[b]
\centering
\vspace{-6mm}\newcommand{\wid}{0.23\textwidth}
\subfloat{\includegraphics[width=\wid]{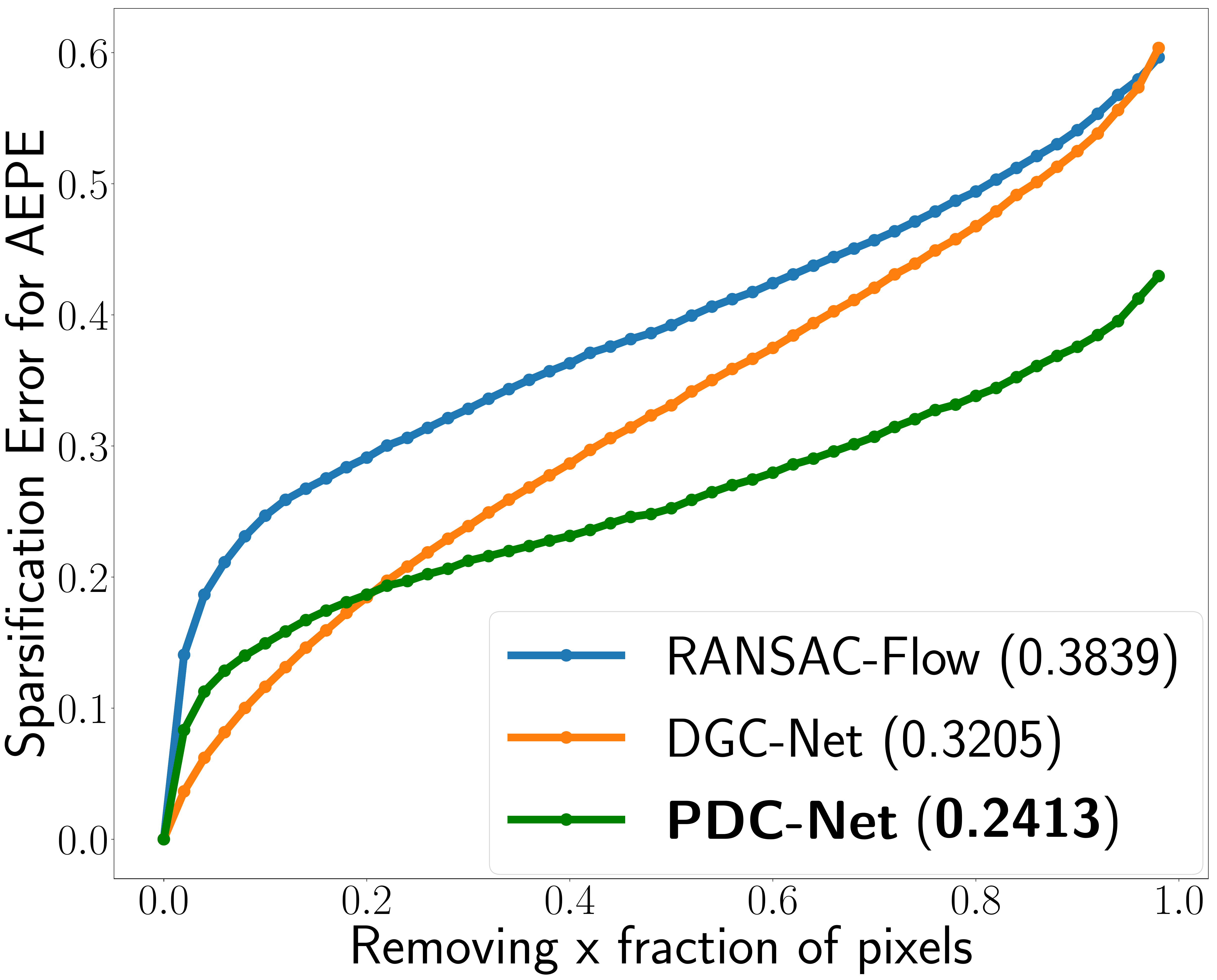}}~%
\subfloat{\includegraphics[width=\wid]{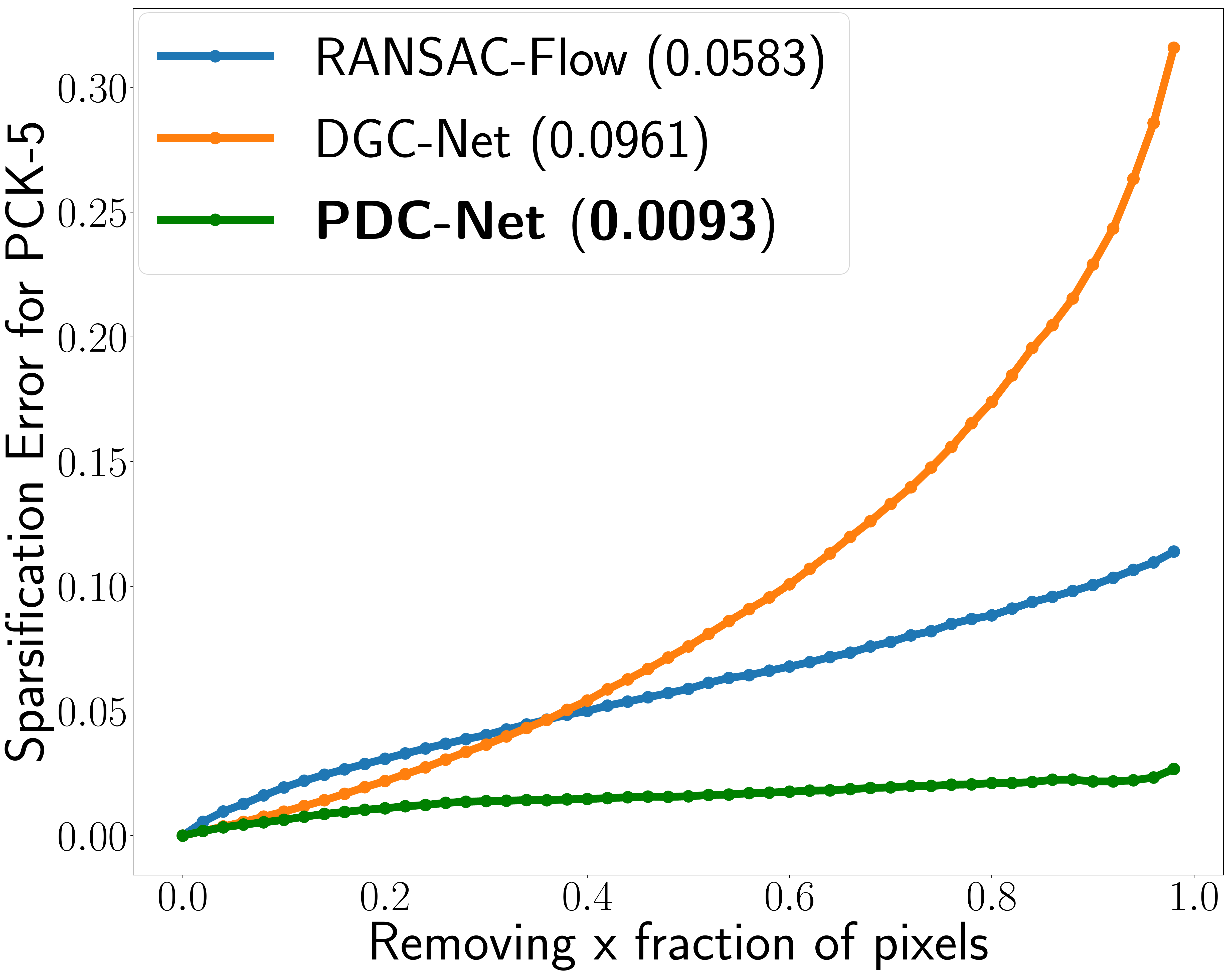}}~%
\caption{Sparsification Error plots for AEPE (left) and PCK-5 (right) on MegaDepth.  Smaller AUSE (in parenthesis) is better. }
\label{fig:sparsification}
\end{figure}

\subsection{Uncertainty Estimation}
\label{subsec:uncertainty-est}

Next, we evaluate our uncertainty estimation. To assess the quality of the uncertainty estimates, we rely on Sparsification Error plots, in line with~\cite{Aodha2013LearningAC,Ilg2017a, ProbFlow}. 
The pixels having the highest uncertainty are progressively removed and the AEPE or PCK of the remaining pixels is calculated, which results in the Sparsification curve.
The Error curve is constructed by subtracting the Sparsification to the Oracle, for which the AEPE and PCK are calculated when the pixels are ranked according to the ground-truth error. As evaluation metric, we use the Area Under the Sparsification Error curve (AUSE).
In Fig.~\ref{fig:sparsification}, we compare the Sparsification Error plots on MegaDepth, of our PDC-Net with other dense methods providing a confidence estimation, namely DGC-Net~\cite{Melekhov2019} and RANSAC-Flow~\cite{RANSAC-flow}. Our probabilistic method PDC-Net produces uncertainty maps that much better correspond to the true errors.

\subsection{Pose and 3D Estimation}
\label{subsec:down-stream-tasks}

Finally, to show the joint performance of our flow and uncertainty prediction, we evaluate our approach for pose estimation. This application has traditionally been dominated by sparse matching methods. 

\parsection{Pose estimation} Given a pair of images showing different view-points of the same scene, two-view geometry estimation aims at recovering their relative pose. We follow the standard set-up of~\cite{OANet} and evaluate on 4 scenes of the \textbf{YFCC100M} dataset~\cite{YFCC}, each comprising 1000 image pairs. As evaluation metrics, we use mAP for different thresholds on the angular deviation between ground truth and predicted vectors for both rotation and translation. 
Results are presented in Tab.~\ref{tab:YCCM}. Our approach PDC-Net outperforms the recent D2D~\cite{D2D} and obtains very similar results than RANSAC-Flow, while being $12.2$ times faster. 
With our multi-scale (MS) strategy, PDC-Net outperforms RANSAC-Flow while being $3.6$ times faster. Note that RANSAC-Flow employs its own MS strategy, using additional off-the-shelf features, which are exhaustively matched with nearest neighbor criteria~\cite{SIFT}. In comparison, our proposed MS is a simpler, faster and more unified approach.  
We also note that RANSAC-Flow relies on a semantic segmentation network to better filter unreliable correspondences, in \eg sky.  Without this segmentation, the performance is drastically reduced. In contrast, our approach can directly estimate highly robust and generalizable confidence maps, without the need for additional network components. The confidence masks of RANSAC-Flow and our approach are visually compared in Fig. \ref{fig:arch-visual-sep-pertur}-\ref{fig:arch-visual-ransac}. 

\parsection{Extension to 3D reconstruction} We also qualitatively show the usability of our approach for dense 3D reconstruction. We compute dense correspondences between day-time images of the Aachen city from the Visual Localization benchmark~\cite{SattlerMTTHSSOP18, SattlerWLK12}. Accurate matches are then selected by thresholding our confidence map, and fed to COLMAP~\cite{SchonbergerF16} to build a 3D point-cloud. It is visualized in Fig. \ref{fig:aachen}.

\begin{table}[b]
\centering
\vspace{-3.5mm}\resizebox{\columnwidth}{!}{%
\begin{tabular}{lccc|c}
\toprule
             &  mAP @5\textdegree &  mAP @10\textdegree & mAP @20\textdegree  & Run-time (s) \\ \midrule
Superpoint~\cite{superpoint} & 30.50 & 50.83 & 67.85 & - \\ 
SIFT~\cite{SIFT} & 46.83 & 68.03 & 80.58 & - \\\midrule          
D2D~\cite{D2D} & 55.58 & 66.79 & - & - \\
RANSAC-Flow (MS+SegNet)~\cite{RANSAC-flow} & 64.88 & 73.31 & 81.56  & 9.06 \\
RANSAC-Flow (MS)~\cite{RANSAC-flow} & 31.25 & 38.76 & 47.36 & 8.99 \\
\textbf{PDC-Net} & 63.90 & 73.00 & 81.22 & \textbf{0.74} \\
\textbf{PDC-Net} (MS) & \textbf{65.18} & \textbf{74.21} & \textbf{82.42} & 2.55 \\
\bottomrule
\end{tabular}%
}\vspace{1mm}\caption{Two-view geometry estimation on YFCC100M~\cite{YFCC}.}
\label{tab:YCCM}
\end{table}

\subsection{Ablation study}
\label{subsec:ablation-study}

Here, we perform a detailed analysis of our approach in Tab.~\ref{tab:ablation}. As baseline, we use a simplified version of GLU-Net, called BaseNet~\cite{GLUNet}. It is a three-level pyramidal network predicting the flow between an image pair. 
Our probabilistic approach integrated in this smaller architecture results in PDC-Net-s. 
All methods are trained using only the first-stage training, described in Sec.~\ref{subsec:arch}.

\parsection{Probabilistic model (Tab.~\ref{tab:ablation},~top)} 
We first compare BaseNet to our approach PDC-Net-s, modelling the flow distribution with a constrained mixture of Laplace (Sec.~\ref{sec:constained-mixture}). On both KITTI-2015 and MegaDepth, our approach brings a significant improvement in terms of flow metrics. Note also that performing pose estimation by taking all correspondences (BaseNet) performs very poorly, which demonstrates the need for robust uncertainty estimation. 
While an unconstrained mixture of Laplace already drastically improves upon the single Laplace component, the permutation invariance of the unconstrained mixture confuses the network, which results in poor uncertainty estimates (high AUSE). Constraining the mixture instead results in better metrics for both the flow and the uncertainty. 

\parsection{Uncertainty architecture (Tab.~\ref{tab:ablation}, middle)}  
While the compared uncertainty decoder architectures achieve similar quality in flow prediction, they provide notable differences in uncertainty estimation. 
Only using the correlation uncertainty module leads to the best results on YFCC100M, since the module enables to efficiently discard unreliable matching regions, in particular compared to the common decoder approach.
However, this module alone does not take into account motion boundaries. This leads to poor AUSE on KITTI-2015, which contains independently moving objects. Our final architecture (Fig.~\ref{fig:arch}), additionally integrating the mean flow into the uncertainty estimation, offers the best compromise.

\parsection{Perturbation data (Tab.~\ref{tab:ablation},~bottom)} 
While introducing the perturbations does not help the flow prediction for BaseNet, it provides  significant improvements in uncertainty \emph{and} in flow performance for our PDC-Net-s. This emphasizes that the improvement of the uncertainty estimates originating from introducing the perturbations, also leads to improved and more generalizable flow predictions.

\begin{table}[t]
\centering
\resizebox{0.48\textwidth}{!}{%
\begin{tabular}{lccc|ccc|cc} 
\toprule
& \multicolumn{3}{c}{\textbf{KITTI-2015}} & \multicolumn{3}{c}{\textbf{MegaDepth}} & \multicolumn{2}{c}{\textbf{YFCC100M}}\\
  & AEPE  & F1 (\%)  & AUSE & PCK-1 (\%)  & PCK-5 (\%) & AUSE & mAP @5\textdegree & mAP @10\textdegree  \\ \midrule
BaseNet (L1-loss) & 7.51 & 37.19 & - & 20.00 & 60.00 & - &  15.58 & 24.00 \\
Single Laplace & 6.86 & 34.27 & 0.220 & 27.45 & 62.24 & 0.210 & 26.95 & 37.10 \\
Unconstrained Mixture & 6.60 & 32.54 & 0.670 & 30.18 & 66.24 & 0.433 & 31.18 & 42.55 \\
Constrained Mixture (PDC-Net-s) & \textbf{6.66} & \textbf{32.32} & \textbf{0.205} & \textbf{32.51} & \textbf{66.50} & \textbf{0.210} & \textbf{33.77} & \textbf{45.17}  \\
\midrule

Commun Dec.  & 6.41 & 32.03 & \textbf{0.171} & 31.93 & \textbf{67.34} & 0.213 & 31.13  & 42.21 \\
Corr unc. module & \textbf{6.32} & \textbf{31.12} & 0.418 & 31.97 & 66.80 & 0.278 & \textbf{33.95} & \textbf{45.44} \\
Unc. Dec. (Fig \ref{fig:arch}) (PDC-Net-s) & 6.66 & 32.32 & 0.205 & \textbf{32.51} & 66.50 & \textbf{0.210} & 33.77 & 45.17  \\ \midrule 

BaseNet \textbf{w/o} Perturbations & 7.21 & 37.35 & - &  20.74 & 59.35 &  - & 15.15 & 23.88 \\
BaseNet \textbf{w} Perturbations  & 7.51 & 37.19 & - & 20.00 & 60.00 & - &  15.58 & 24.00 \\
PDC-Net-s \textbf{w/o} Perturbations  & 7.15 & 35.28 & 0.256 &  31.53 & 65.03 & 0.219 & 32.50 & 43.17 \\
PDC-Net-s \textbf{w} Perturbations & \textbf{6.66} & \textbf{32.32} & \textbf{0.205} & \textbf{32.51} & \textbf{66.50} & \textbf{0.210} & \textbf{33.77} & \textbf{45.17} \\
\bottomrule
\end{tabular}%
}\vspace{1mm}
\caption{Ablation study. In the top part, different probabilistic models are compared (Sec.~\ref{subsec:proba-model}-\ref{sec:constained-mixture}). In the middle part, a constrained Mixture is used, and different architectures for uncertainty estimation are compared (Sec.~\ref{sec:uncertainty-arch}). In the bottom part, we analyze the impact of our training data with perturbations (Sec.~\ref{subsec:perturbed-data}).}\vspace{-5mm}
\label{tab:ablation}
\end{table}
\section{Conclusion}

We propose a probabilistic deep network for estimating the dense image-to-image correspondences and associated confidence estimate. Specifically, we train the network to predict the parameters of the conditional probability density of the flow, which we model with a constrained mixture of Laplace distributions. Moreover, we introduce an architecture and improved self-supervised training strategy, designed for \emph{robust and generalizable} uncertainty prediction.
Our approach PDC-Net sets a new state-of-the-art on multiple geometric matching and optical flow datasets. It also outperforms dense matching methods on pose estimation. 

\parsection{Acknowledgements}
This work was supported by the ETH Z\"urich Fund (OK), a Huawei Gift, Huawei Technologies Oy (Finland), Amazon AWS, and an Nvidia GPU grant.

{\small
\bibliographystyle{ieee_fullname}
\bibliography{biblio}
}

\clearpage
\newpage
\appendix
\begin{center}
	\textbf{\Large Appendix}
\end{center}

In this appendix, we first give a detailed derivation of our probabilistic model as a constrained mixture of Laplace distributions in Sec.~\ref{sec-sup:proba}. In Sec.~\ref{sec-sup:training}, we then derive our probabilistic training loss and explain our training procedure in more depth.  We subsequently follow by providing additional information about the architecture of our proposed networks as well as implementation details in Sec.~\ref{sec-sub:arch-details}. 
In Sec.~\ref{sec-sup:details-evaluation}, we extensively explain the evaluation datasets and set-up.  Then, we present more detailed quantitative and qualitative results in Sec.~\ref{sec-sup:results}.
Finally, we perform detailed ablative experiments in Sec.~\ref{sec-sup:ablation}.

\section{Detailed derivation of probabilistic model}
\label{sec-sup:proba}

Here we provide the details of the derivation of our uncertainty estimate. 

\parsection{Probabilistic formulation} We model the flow estimation as a probabilistic regression with a constrained mixture density of Laplacian distributions (Sec.~\ref{sec:constained-mixture} of the main paper). Our mixture model, corresponding to equation \eqref{eq:mixture} of the main paper, is expressed as,
\begin{equation}
\label{eq:mixture-sup}
p\left(y | \varphi \right) = \sum_{m=1}^{M} \alpha_{m} \mathcal{L}(y| \mu, \sigma^2_m)
\end{equation}
where, for each component $m$, the bi-variate Laplace distribution $\mathcal{L}(y| \mu, \sigma^2_m)$ is computed as the product of two independent uni-variate Laplace distributions, such as, 
\begin{subequations}
\label{eq:simple-laplace}
\begin{align}
\mathcal{L}(y| \mu, \sigma^2_m) &= \mathcal{L}(u,v| \mu_u, \mu_v, \sigma^2_u, \sigma^2_v) \\
&= \mathcal{L}(u| \mu_u, \sigma^2_u). \mathcal{L}(v| \mu_v, \sigma^2_v) \\
&= \frac{1}{\sqrt{2 \sigma_u^2}} e^{-\sqrt{\frac{2}{\sigma_u^2}}|u-\mu_u|} .  \nonumber \\
& \hspace{5mm} \frac{1}{\sqrt{2 \sigma_v^2}}e^{-\sqrt{\frac{2}{\sigma_v^2}}|v-\mu_v|}
\end{align}
\end{subequations}
where $\mu = [\mu_u, \mu_v]^T \in \mathbb{R}^2$ and  $\sigma^2_m = [\sigma^2_u, \sigma^2_v]^T \in \mathbb{R}^2$ are respectively the mean and the variance parameters of the distribution $\mathcal{L}(y| \mu, \sigma^2_m)$. In this work, we additionally define equal variances in both flow directions, such that $\sigma^2_m = \sigma^2_u = \sigma^2_v \in \mathbb{R}$. As a result, equation \eqref{eq:simple-laplace} simplifies, and when inserting into \eqref{eq:mixture-sup}, we obtain,
\begin{equation}
\label{eq:mixture-sup-final}
p\left(y | \varphi \right) =\sum_{m=1}^{M} \alpha_{m} \frac{1}{2 \sigma_m^2} e^{-\sqrt{\frac{2}{\sigma_m^2}}|y-\mu|_1}.
\end{equation}

\begin{figure}[t]
\centering
(A) MegaDepth \\
\includegraphics[width=0.47\textwidth]{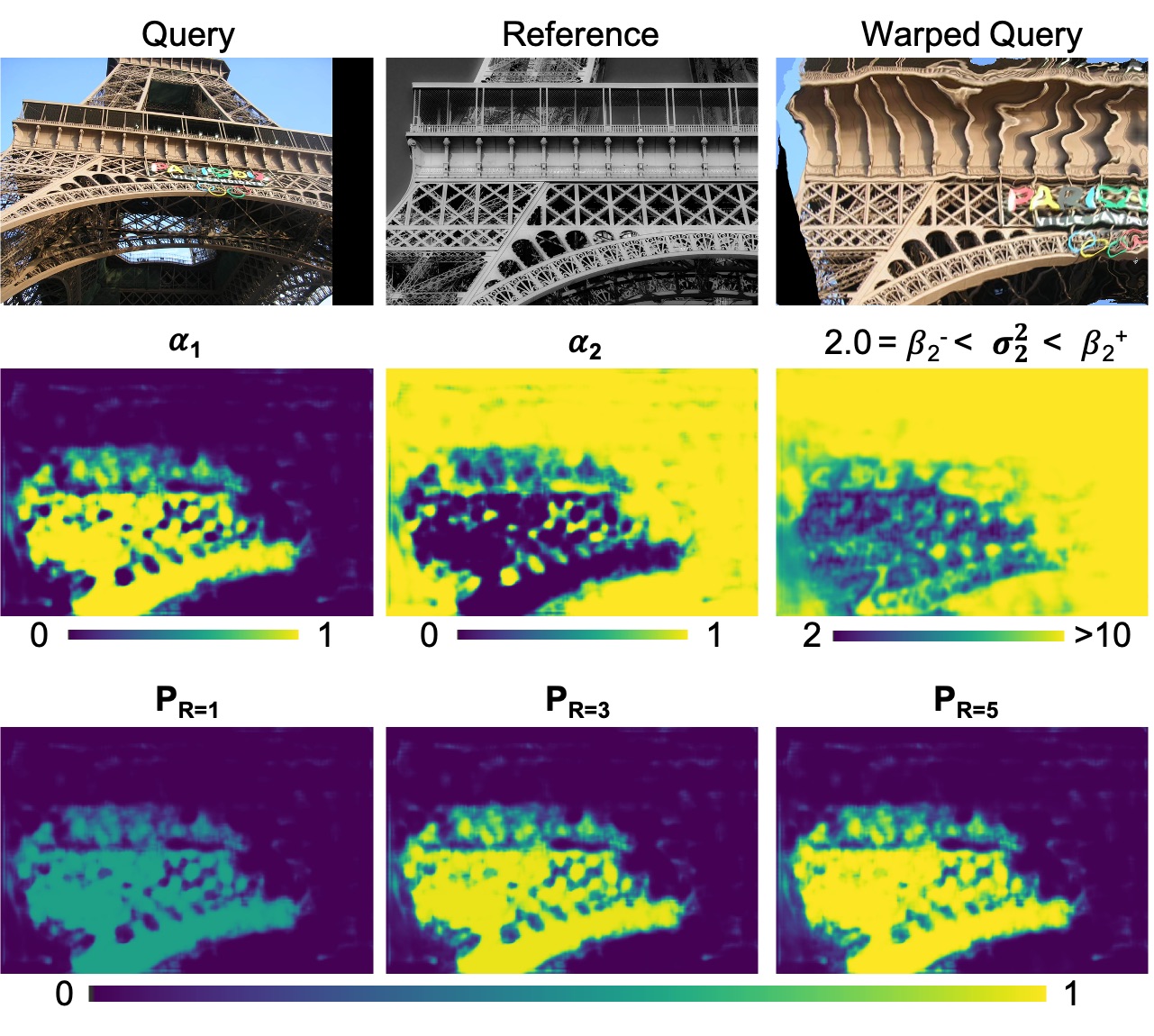} \\
(B) KITTI-2015 \\
\includegraphics[width=0.47\textwidth]{ 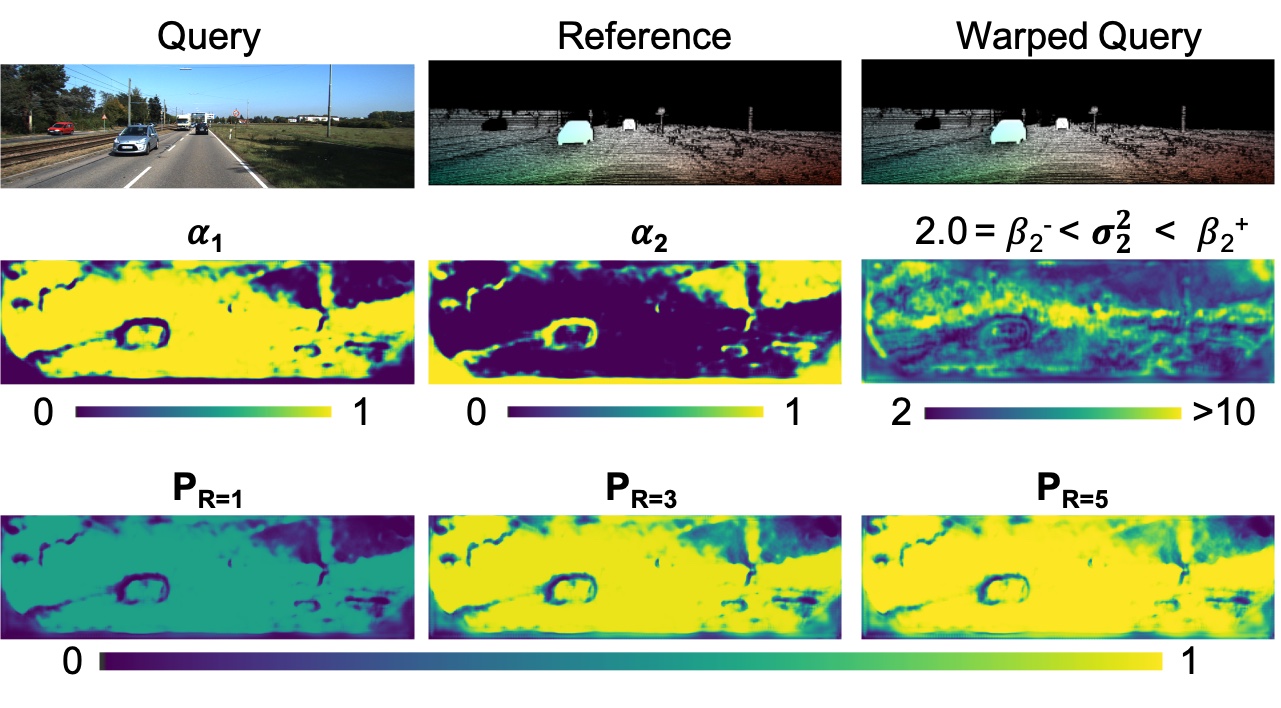} \\
\caption{Visualization of the mixture parameters $(\alpha_m )_{m=1}^M$ and $\sigma^2_2$ predicted by our final network PDC-Net, on multiple image pairs. PDC-Net has $M=2$ Laplace components and here, we do not represent the scale parameter $\sigma^2_1$, since it is fixed as $\sigma^2_1 = 1.0$. We also show the resulting confidence maps $P_R$ for multiple R. }\vspace{-4mm}
\label{fig:conf-map}
\end{figure}

\parsection{Confidence estimation} Our network $\Phi$ thus outputs, for each pixel location, the parameters of the predictive distribution, \ie the mean flow $\mu$ along with the variance $\sigma^2_m$ and weight $\alpha_m$ of each component, as $\big( \mu, (\alpha_m )_{m=1}^M, ( \sigma^2_m  )_{m=1}^M  \big) = \varphi(X; \theta)$. However, we aim at obtaining a \emph{single} confidence value to represent the relibiability of the estimated flow vector $\mu$.
As a final confidence measure,  we thus compute the probability $P_R$ of the true flow being within a radius $R$ of the estimated mean flow vector $\mu$. This is expressed as,
\begin{subequations}
\begin{align}
    P_R & = P(\|y - \mu\|_\infty < R) \\
    & = \int_{\{y\in \reals^2: \|y - \mu\|_\infty < R\}}\! p(y|\varphi) dy \\
    &= \sum_{m} \alpha_m \int_{\mu_u-R}^{\mu_u+R}
    \frac{1}{\sqrt{2} \sigma_m} e^{-\sqrt{2}\frac{|u-\mu_u|}{\sigma_m}} du   \nonumber \\
      &    \hspace{15mm}   \int_{\mu_v-R}^{\mu_v+R}
    \frac{1}{\sqrt{2} \sigma_m} e^{-\sqrt{2}\frac{|v-\mu_v|}{\sigma_m}}dv \\
    &= \sum_{m} \alpha_m \left[1-\exp (-\sqrt{2}\frac{R}{\sigma_m}) \right]^2
\end{align}
\end{subequations}
where we have here expressed $P_R$ with the standard deviation parameters $\sigma_m$ instead of the variance parameters $\sigma_m^2$ for ease of notation. 
This confidence measure is used to identify the accurate matches by thesholding $P_R$. In Fig~\ref{fig:conf-map}, we visualize the estimated mixture parameters $(\alpha_m )_{m=1}^M, ( 
\sigma^2_m  )_{m=1}^M$, and the resulting confidence map $P_R$ for multiple image pair examples.

\section{Training details}
\label{sec-sup:training}

In this section, we derive the numerically stable Negative Log-Likelihood loss, used for training our network PDC-Net. We also describe in details the employed training datasets. 

\subsection{Training loss}
\label{sec-sup:training-loss}

Similar to conventional approaches, probabilistic methods are generally trained using a set of \emph{iid}.\
image pairs $\mathcal{D} = \left \{X^{(n)}, Y^{(n)} \right \}_{n=1}^N$.
The negative log-likelihood provides a general framework for fitting a distribution to the training dataset as,
\begin{subequations}
\label{eq:nll-sup}
\begin{align}
L(\theta; \mathcal{D}) &=  - \frac{1}{N} \sum_{n=1}^{N} \log p\left(Y^{(n)} | \Phi(X^{(n)}; \theta) \right)  \\
&= - \frac{1}{N} \sum_{n=1}^{N} \sum_{ij} \log p\big(y^{(n)}_{ij} | \varphi_{ij}(X^{(n)};\theta)\big) 
\end{align}
\end{subequations}
Inserting \eqref{eq:mixture-sup-final} into \eqref{eq:nll-sup}, we obtain for the last term the following expression,
\begin{subequations}
\label{eq:p-sup}
\begin{align}
L_{ij} &= -\log p\big(y^{(n)}_{ij} | \varphi_{ij}(X^{(n)};\theta)\big) \\
&= -\log \left ( \sum_{m=1}^{M} \alpha_{m} \frac{1}{2 \sigma_m^2} e^{  -\sqrt{\frac{2}{\sigma_m^2}}|y-\mu|_1  }   \right) \\
&=  -\log \left(   \sum_{m=1}^{M} \frac{e^{\tilde{\alpha}_m}}{\sum_{m=1}^{M}e^{\tilde{\alpha}_m}} \frac{1}{2 \sigma_m^2} e^{   -\sqrt{\frac{2}{\sigma_m^2}}|y-\mu|_1  } \right) \\
&=\log  \left( \sum_{m=1}^{M}e^{\tilde{\alpha}_m} \right) \nonumber  \\ 
& \;\;\;\;\;  -\log \left(  \sum_{m=1}^{M} 
e^{\tilde{\alpha}_m}\frac{1}{2 \sigma_m^2} e^{  -\sqrt{\frac{2}{\sigma_m^2}}|y-\mu|_1  } \right) \\
&= \log   \left( \sum_{m=1}^{M}e^{\tilde{\alpha}_m}  \right) \nonumber \\ & 
\;\;\;\;\; -\log \left(\sum_{m=1}^{M} e^{\tilde{\alpha}_m - \log(2) - s_m -\sqrt{2}e^{-\frac{1}{2}s_m}.|y-\mu|_1}  \right)   
\end{align}
\end{subequations}
where $s_m = log(\sigma^2_m)$. Indeed, in practise, to avoid division by zero, we use $s_{m} = \log(\sigma^2_m)$ for all components $m \in \left \{0, .., M \right \}$ of the mixture density model. For the implementation of the loss, we use a numerically stable logsumexp function.

With a simple regression loss such as the L1 loss, the large errors represented by the heavy tail of the distribution in Fig. \ref{fig:distribution} of the main paper have a disproportionately large impact on the loss, preventing the network from focusing on the more accurate predictions. On the contrary, the loss \eqref{eq:p-sup} enables to down-weight the contribution of these examples by predicting a high variance parameter for them. Modelling the flow estimation as a conditional predictive distribution thus improves the accuracy of the estimated flow itself.

\subsection{Training datasets}

Due to the limited amount of available real correspondence data, most matching methods resort to self-supervised training, relying on synthetic image warps generated automatically. We here provide details on the synthetic dataset that we use for self-supervised training, as well as additional information on the implementation of the perturbation data (Sec. \ref{subsec:perturbed-data} of the main paper). Finally, we also describe the generation of the sparse ground-truth correspondence data from the MegaDepth dataset~\cite{megadepth}. 

\parsection{Base synthetic dataset}  For our base synthetic dataset, we use the same data as in~\cite{GOCor}. Specifically, pairs of images are created by warping a collection of images from the DPED~\cite{Ignatov2017}, CityScapes~\cite{Cordts2016} and ADE-20K~\cite{Zhou2019} datasets, according to synthetic affine, TPS and homography transformations. The transformation parameters are the ones originally used in DGC-Net~\cite{Melekhov2019}. 

These image pairs are further augmented with additional random independently moving objects. To do so, the objects are sampled from the COCO dataset~\cite{coco}, and inserted on top of the images of the synthetic data using their segmentation masks. To generate motion, we randomly sample affine transformation parameters for the foreground objects, which are independent of the background transformations. This can be interpreted as both the camera and the objects moving independently of each other. The final synthetic flow is composed of the object motion flow field at the location of the moving object in the reference image, or the background flow field otherwise. 
It results in 40K image pairs, cropped at resolution $520 \times 520$.

\begin{figure}[t]
\centering
\includegraphics[width=0.49\textwidth]{ 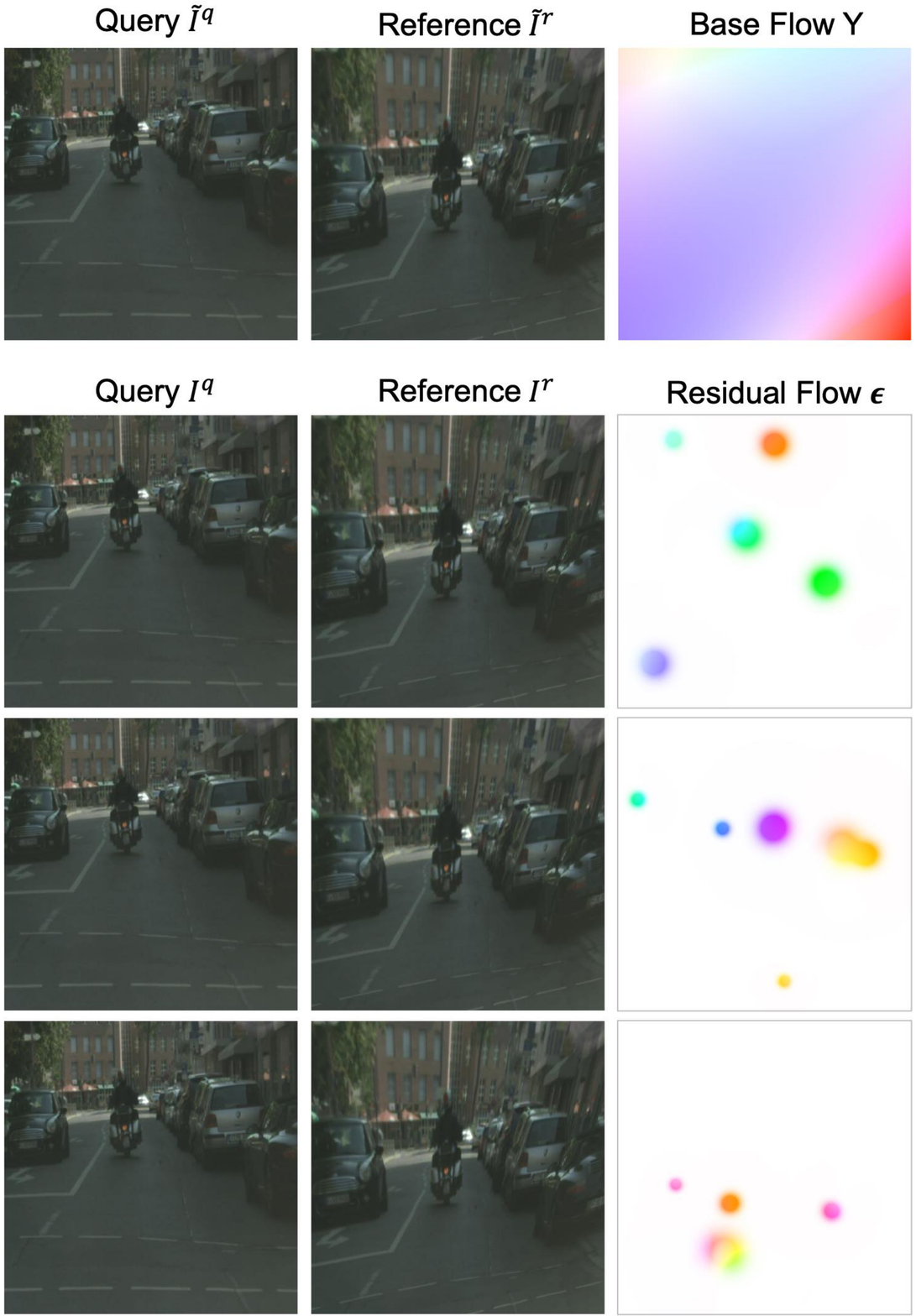}
\caption{Visualization of our perturbations applied to a pair of reference and query images (Sec.~\ref{subsec:perturbed-data} of the main paper). }\vspace{-2mm}
\label{fig:pertu}
\end{figure}

\parsection{Perturbation data for robust uncertainty estimation} Even with independently moving objects, the network still learns to primarily rely on interpolation when estimating the flow field and corresponding uncertainty map relating an image pair. We here describe in more details our data generation strategy for more robust uncertainty prediction. From a base flow field $Y \in \reals^{H \times W \times 2}$ relating a reference image $\tilde{I}^r \in \reals^{H \times W \times 3}$ to a query image $\tilde{I}^q \in \reals^{H \times W \times 3}$, we introduce a residual flow $\epsilon = \sum_i \varepsilon_i$, by adding small local perturbations $\varepsilon_i \in \reals^{H \times W \times 2}$.
More specifically, we create the residual flow by first generating an elastic deformation motion field $E$ on a dense grid of dimension $H \times W$, as described in~\cite{Simard2003}. Since we only want to include perturbations in multiple small regions, we generate binary masks $S_i \in \mathbb{R}^{H \times W \times 2}$, each delimiting the area on which to apply one local perturbation $\varepsilon_i$.  The final residual flow (perturbations) thus take the form of $\epsilon = \sum_i \varepsilon_i$, where $\varepsilon_i = E \cdot S_i$. 
Finally, the query image $I^q = \tilde{I}^q$ is left unchanged  while the reference $I^r$ is generated by warping $\tilde{I}^r$ according to the residual flow $\epsilon$, as $I^r(x) = \tilde{I^r}(x+\epsilon(x))$. The final perturbed flow map $Y$ between $I^r$ and $I^q$ is achieved by composing the base flow $\tilde{Y}$ with the residual flow $\epsilon$, as $Y(x) = \tilde{Y}(x+\epsilon(x)) + \epsilon(x)$.

In practise, for the elastic deformation field $E$, we use the implementation of~\cite{info11020125}. The masks $S_i$ should be between 0 and 1 and offer a smooth transition between the two, so that the perturbations appear smoothly. To create each mask $S_i$, we thus generate a 2D Gaussian centered at a random location and with a random standard deviation (up to a certain value) on a dense grid of size $H \times W$. It is then scaled to 2.0 and clipped to 1.0, to obtain a smooth regions equal to 1.0 where the perturbation will be applied, and transition regions on all sides from 1.0 to 0.0. 

In Fig.~\ref{fig:pertu}, we show examples of generated residual flows and their corresponding perturbed reference $I^r$, for a particular base flow $Y$, and query $\tilde{I}^r$ and reference $\tilde{I}^q$ images. As one can see, for human eye, it is almost impossible to detect the presence of the perturbations on the perturbed reference $I^r$. This will enable to "fool" the network in homogeneous regions, such as the road in the figure example, thus forcing it to predict high uncertainty in regions where it cannot identify them.

\parsection{MegaDepth training} To generate the training pairs with sparse ground-truth, we adapt the generation protocol of D2-Net~\cite{Dusmanu2019CVPR}. Specifically, we use the MegaDepth dataset, consisting of 196 different scenes reconstructed from 1.070.468 internet photos using COLMAP~\cite{SchonbergerF16}. The camera intrinsics and extrinsics as well as depth maps from Multi-View Stereo are provided by the authors for 102.681 images.

For training, we use 150 scenes and sample up to 500 random images with an overlap ratio of at least 30\% in the sparse SfM point cloud. For each pair, all points of the second image with depth information are projected into the first image. A depth-check with respect to the depth map of the first image is also run to remove occluded pixels. It results in around 58.000 training pairs, which we resized so that their largest dimension is 520. Note that we use the same set of training pairs at each training iteration. 
For the validation dataset, we sample up to 100 image pairs from 25 different scenes, leading to  approximately 1800 image pairs.

During the second stage of training, we found it crucial to train on \emph{both} the synthetic dataset with perturbations and the sparse data from MegaDepth. Training solely on the sparse correspondences resulted in less reliable uncertainty estimates.

\section{Architecture details}
\label{sec-sub:arch-details}

In this section, we first describe the architecture of our proposed uncertainty decoder (Sec. \ref{sec:uncertainty-arch} of the main paper). We then give additional details about our proposed final architecture PDC-Net and its corresponding baseline GLU-Net-GOCor*. We also describe the architecture of BaseNet and its probabilistic derivatives, employed for the ablation study. Finally, we share all training details and hyper-parameters.

\subsection{Architecture of the uncertainty decoder}
\label{subsec-sup:unc-dec}

\parsection{Correlation uncertainty module} We first describe the architecture of our Correlation Uncertainty Module $U_{\theta}$ (Sec. \ref{sec:uncertainty-arch} of the main paper). The correlation uncertainty module processes each 2D slice $C_{ij\cdot \cdot}$ of the correspondence volume $C$ independently. 
More practically, from the correspondence volume tensor $C \in \mathbb{R}^ {b \times h \times w \times (d \times d)}$, where b indicates the batch dimension, we move the spatial dimensions $h \times w$ into the batch dimension and we apply multiple convolutions in the displacement dimensions $d \times d$, \ie on a tensor of shape $(b \times h \times w) \times d \times d \times 1$. By applying the strided convolutions, the spatial dimension is gradually decreased, resulting in an uncertainty representation $u \in \mathbb{R}^{(b \times h \times w) \times 1 \times 1 \times n}$, where $n$ denotes the number of channels. $u$ is subsequently rearranged, and after dropping the batch dimension, the outputted uncertainty tensor is $u \in \mathbb{R}^{h \times w \times n}$.

Note that while this is explained for a local correlation, the same applies for a global correlation except that the displacement dimensions correspond to $h \times w$. 
In Tab.~\ref{tab:arch-local-disp}-~\ref{tab:arch-global-disp}, we present the architecture of the convolution layers applied on the displacements dimensions, for a local correlation with search radius 4 and for a global correlation applied at dimension $ h \times w = 16 \times 16$, respectively.

\begin{table}[b]
\centering
\resizebox{0.49\textwidth}{!}{%
\begin{tabular}{l|l|l}
\toprule
Inputs & Convolutions & Output size \\ \midrule
C; $(b \times h \times w) \times 9 \times 9 \times 1$ & $conv_0$, $K=(3 \times 3)$, s=1, p=0 &  $(b \times h \times w) \times 7 \times 7 \times 32$ \\ \midrule
$conv_0$; $(b \times h \times w) \times 7 \times 7 \times 32$ & $conv_1$, $K=(3 \times 3)$, s=1, p=0 &  $(b \times h \times w) \times 5 \times 5 \times 32$ \\ \midrule
$conv_1$; $(b \times h \times w) \times 5 \times 5 \times 32$ & $conv_2$, $K=(3 \times 3)$, s=1, p=0 &  $(b \times h \times w) \times 3 \times 3 \times 16$ \\ \midrule
$conv_2$; $(b \times h \times w) \times 3 \times 3 \times 16$ & $conv_3$, $K=(3 \times 3)$, s=1, p=0 &  $(b \times h \times w) \times 1 \times 1 \times n$ \\ 
\bottomrule
\end{tabular}%
}\vspace{1mm}\caption{Architecture of the correlation uncertainty module for a local correlation, with a displacement radius of 4. Implicit are the BatchNorm and ReLU operations that follow each convolution, except for the last one. K refers to kernel size, s is the used stride and p the padding. }
\label{tab:arch-local-disp}
\end{table}

\begin{table}[b]
\centering
\resizebox{0.49\textwidth}{!}{%
\begin{tabular}{l|l|l}
\toprule
Inputs & Convolutions & Output size \\ \midrule
C $(b \times h \times w) \times 16 \times 16 \times 1$ & $conv_0$;  $K=(3 \times 3)$, s=1, p=0  &  $(b \times h \times w) \times 14 \times 14 \times 32$ \\ \midrule
$conv_0$; $(b \times h \times w) \times 14 \times 14 \times 32$ &  \begin{tabular}[c]{@{}l@{}}$3 \times 3$ max pool, s=2 \\ $conv_1$, $K=(3 \times 3)$, s=1, p=0\end{tabular}  & $(b \times h \times w) \times 5 \times 5 \times 32$ \\ \midrule

$conv_1$; $(b \times h \times w) \times 5 \times 5 \times 32$ & $conv_2$, $K=(3 \times 3)$, s=1, p=0 &  $(b \times h \times w) \times 3 \times 3 \times 16$ \\ \midrule
$conv_2$; $(b \times h \times w) \times 3 \times 3 \times 16$ & $conv_3$,  $K=(3 \times 3)$, s=1, p=0 &  $(b \times h \times w) \times 1 \times 1 \times n$ \\ 
\bottomrule
\end{tabular}%
}\vspace{1mm}\caption{Architecture of the correlation uncertainty module for a global correlation, constructed at resolution $16 \times 16$. Implicit are the BatchNorm and ReLU operations that follow each convolution, except for the last one. K refers to kernel size, s is the used stride and p the padding. }
\label{tab:arch-global-disp}
\end{table}

\parsection{Uncertainty predictor} We then give additional details of the Uncertainty Predictor, that we denote $Q_{\theta}$ (Sec. \ref{sec:uncertainty-arch} of the main paper). 
The uncertainty predictor takes the flow field $Y \in \reals^{h \times w \times 2}$ outputted from the flow decoder, along with the output $u \in \reals^{h \times w \times n}$ of the correlation uncertainty module $U_{\theta}$. In a multi-scale architecture, it additionally takes as input the estimated flow field and predicted uncertainty components from the previous level. At level $l$, for each pixel location $(i,j)$, this is expressed as:
\begin{equation}
\big((\tilde{\alpha}_m )_{m=1}^M, ( h_m  )_{m=1}^M  \big)^l = Q_{\theta} \big( Y^l; u^l;  \Phi^{l-1} \big)_{ij}
\end{equation}
where $\tilde{\alpha}_m$ refers to the output of the uncertainty predictor, which is then passed through a SoftMax layer to obtain the final weights $\alpha_m$. $\sigma^2_m$ is obtained from $h_m$ according to constraint equation \eqref{eq:constraint} of the main paper. 

In practise, we have found that instead of feeding the flow field $Y \in \reals^{h \times w \times 2}$ outputted from the flow decoder to the uncertainty predictor, using the second last layer from the flow decoder leads to slightly better results. This is because the second last layer from the flow decoder has larger channel size, and therefore encodes more information about the estimated flow.

Architecture-wise, the uncertainty predictor $Q_{\theta}$ consists of 3 convolutional layers. The numbers of feature channels at each convolution layers are respectively 32, 16 and $2M$ and the spatial kernel of each convolution is $3 \times 3$ with stride of 1 and padding 1. The first two layers are followed by a batch-normalization layer with a leaky-Relu non linearity. The final output of the uncertainty predictor is the result of a linear 2D convolution, without any activation.

\subsection{Architecture of PDC-Net}
\label{subsec-sup:pdcnet}

We use GLU-Net-GOCor~\cite{GLUNet, GOCor} as our base architecture, predicting the dense flow field relating a pair of images. It is a 4 level pyramidal network, using a VGG feature backbone. It is composed of two sub-networks, L-Net and H-Net which act at two different resolutions. 
The L-Net takes as input rescaled images to $H_L \times W_L= 256 \times 256$ and process them with a global GOCor module followed by a local GOCor module. The resulting flow is then upsampled to the lowest resolution of the H-Net to serve as initial flow, by warping the query features according to the estimated flow. The H-Net takes input images at unconstrained resolution $H\times W$, and refines the estimated flow with two local GOCor modules. 

For the baseline GLU-Net-GOCor*, we adopt the GLU-Net-GOCor architecture and simply replace the DenseNet connections~\cite{Huang2017} of the flow decoders by standard residual blocks. The mapping decoder is also modified to include residual connections. This drastically reduces the number of weights while not having any impact on performance. As in~\cite{GLUNet, GOCor}, the VGG-16 backbone is initialized to the pre-trained weights on ImageNet.  

From the baseline GLU-Net-GOCor*, we create our probabilistic approach PDC-Net by inserting our uncertainty decoder at each pyramid level. As noted in \ref{subsec-sup:unc-dec}, in practise, we feed the second last layer from the flow decoder to the uncertainty predictor of each pyramid level instead of the predicted flow field. It leads to slightly better results. 
The uncertainty prediction is additionally \emph{propagated from one level to the next}. More specifically, the flow decoder takes as input the uncertainty prediction (all parameters $\Phi$ of the predictive distribution except for the mean flow) of the previous level, in addition to its original inputs (which include the mean flow of the previous level). The uncertainty predictor also takes the uncertainty and the flow estimated at the previous level. 
As explained in Sec. \ref{subsec:arch} of the main paper, we use a constrained mixture with $M=2$ Laplace components. 
The first component is set so that $\sigma^2_1 = 1$, while the second is learned as $ 2 = \beta_2^- \leq \sigma^2_2 \leq \beta_2^+$. Therefore, the uncertainty predictor only estimates $\sigma^2_2$ and $(\alpha_m )_{m=1}^{M=2}$ at each pixel location.
We found that fixing $\sigma^2_1 = \beta_1^- = \beta_1^+ = 1.0$ results in better performance than for example $\beta_1^- = 0.0 < \sigma^2_1 < \beta_1^+=1.0$. Indeed, in the later case, during training, the network focuses primarily on getting the very accurate, and confident, correspondences (corresponding to $\sigma^2_1$) since it can arbitrarily reduce the variance. Generating fewer, but accurate predictions then dominate during training to the expense of other regions. This is effectively alleviated by setting $\sigma^2_1=1.0$, which can be seen as introducing a strict prior on this parameter. 

\subsection{Inference multi-stage}

Here, we provide implementation details for our multi-stage inference strategy (Sec. \ref{sec:geometric-inference} of the main paper). 
After the first network forward pass, we select matches with a corresponding confidence probability $P_{R=1}$ superior to 0.1, for $R=1$.  Since the network estimates the flow field at a quarter of the original image resolution, we use the filtered correspondences at the quarter resolution and scale them to original resolution to be used for homography estimation.  To estimate the homography, we use OpenCV’s findHomography with RANSAC and an inlier threshold of 1 pixel.

\subsection{Inference multi-scale}

We then give additional details about our multi-scale strategy (MS). We extend our two-stage refinement approach (Sec.~\ref{sec:geometric-inference}) by resizing the reference image to different resolutions. Specifically, following~\cite{RANSAC-flow}, we use seven scales: 0.5, 0.88, 1, 1.33, 1.66 and 2.0.
As for the implementation, to avoid obtaining very large input images (for scaling ratio 2 for example), we use the following scheme: we resize the reference image for scaling ratios below 1, keeping the aspect ratio fixed and the query image unchanged. For ratios above 1, we instead resize the query image by one divided by the ratio, while keeping the reference image unchanged. This ensures that the resized images are never larger than the original image dimensions. 
The resulting image pairs are then passed through the network and we fit a homography for each pair, using our predicted flow and uncertainty map. In particular, as in our two-stage inference strategy, we select matches with a corresponding confidence probability $P_{R=1}$ superior to 0.1, for $R=1$, at the estimated flow resolution, \ie at a quarter of the input image resolution. To estimate the homography, we use OpenCV’s findHomography with RANSAC and an inlier threshold of 1 pixel.
From all image pairs with their corresponding scaling ratios, we then select the homography with the highest percentage of inliers, and scale it to the images original resolutions. The original image pair is then coarsely aligned using this homography and from there we follow the same procedure, as explained in Sec.~\ref{sec:geometric-inference}.  

\subsection{Architecture of BaseNet}
\label{arch-basenet}

As baseline to use in our ablation study, we use BaseNet, introduced in~\cite{GOCor} and inspired by GLU-Net~\cite{GLUNet}. It estimates the dense flow field relating an input image pair. 
The network is composed of three pyramid levels and it uses VGG-16~\cite{Chatfield14} as feature extractor backbone. The coarsest level is based on a global correlation layer, followed by a mapping decoder estimating the correspondence map at this resolution. The two next pyramid levels instead rely on local correlation layers. The dense flow field is then estimated with flow decoders, taking as input the correspondence volumes resulting from the local feature correlation layers. Moreover, BaseNet is restricted to a pre-determined input resolution $H_L \times W_L = 256 \times 256$ due to its global correlation at the coarsest pyramid level. It estimates a final flow-field at a quarter of the input resolution $H_L \times W_L$, which needs to be upsampled to original image resolution $H \times W$. The mapping and flow decoders have the same number of layers and parameters as those used for GLU-Net~\cite{GLUNet}. However, here, to reduce the number of weights, we use feed-forward layers instead of DenseNet connections~\cite{Huang2017} for the flow decoders.

We create the different probabilistic versions of BaseNet, presented in the ablation study Tab.~\ref{tab:ablation} of the main paper, by modifying the architecture minimally. Moreover, for the probabilistic versions modeled with a constrained mixture, we use $M=2$ Laplace components. The first component is set so that $\sigma^2_1 = 1$, while the second is learned as $2 = \beta_2^- \leq \sigma^2_2 \leq \beta_2^+ = \infty$. For the network referred to as PDC-Net-s, which also employs our proposed uncertainty architecture (Sec.~\ref{sec:uncertainty-arch} of the main paper), we add our uncertainty decoder at each pyramid layer, in a similar fashion as for our final network PDC-Net. 
We train all networks on the synthetic data with the perturbations, which corresponds to our first training stage (Sec.~\ref{subsec:arch}).

\subsection{Implementation details} 

Since we use pyramidal architectures with $K$ levels, we employ a multi-scale training loss, where the loss at different pyramid levels account for different weights. 
\begin{equation}
\label{eq:multiscale-loss}
    \mathcal{L}(\theta)=\sum_{l=1}^{K} \gamma_{l} L_l +\eta \left\| \theta \right\|\,,
\end{equation}
where  $\gamma_{l}$ are the weights applied to each pyramid level and $L_l$ is the corresponding loss computed at each level, which refers to the L1 loss for the non-probabilistic baselines and the negative log-likelihood loss \eqref{eq:nll-sup} for the probabilistic models, including our approach PDC-Net. The second term of the loss \eqref{eq:multiscale-loss} regularizes the weights of the network. 
Moreover, during the self-supervised training, we do not apply any mask, which means that out-of-view regions (that do not have visible matches) are included in the training loss. Since the image pairs are related by synthetic transformations, these regions do have a correct ground-truth flow value. 
When finetuning on MegaDepth images however, the loss is applied only at the locations of the sparse ground-truth.

For training, we use similar training parameters as in~\cite{GLUNet}. Specifically, as a preprocessing step, the training images are mean-centered and normalized using mean and standard deviation of ImageNet dataset~\cite{Hinton2012}. For all local correlation layers, we employ a search radius $r=4$. 

For the training of BaseNet and its probabilistic derivatives (including PDC-Net-s), which have a pre-determined fixed input image resolution of $(H_L \times W_L = 256 \times 256)$, we use a batch size of 32 and train for 106.250 iterations. We set the initial learning rate to $10^{-2}$ and gradually decrease it by 2 after 56.250, 75.000 and 93.750 iterations. 
The weights in the training loss \eqref{eq:multiscale-loss} are set to be $\gamma_{1}=0.32, \gamma_{2}=0.08, \gamma_{3}=0.02$ and to compute the loss, we down-sample the ground-truth to estimated flow resolution at each pyramid level. 

For GLU-Net-GOCor* and PDC-Net, we down-sample and scale the ground truth from original resolution $H\times W$ to $H_L \times W_L$ in order to obtain the ground truth flow fields for L-Net.
During the first stage of training, \ie on purely synthetic images, we down-sample the ground truth from the base resolution to the different pyramid resolutions without further scaling, so as to obtain the supervision signals at the different levels. During this stage, the weights in the training loss \eqref{eq:multiscale-loss} are set to be $\gamma_{1}=0.32, \gamma_{2}=0.08, \gamma_{3}=0.02, \gamma_{4}=0.01$, which ensures that the loss computed at each pyramid level contributes equally to the final loss \eqref{eq:multiscale-loss}.  
During the second stage of training however, which includes MegaDepth, since the ground-truth is sparse, it is inconvenient to down-sample it to different resolutions. We thus instead up-sample the estimated flow field to the ground-truth resolution and compute the loss at this resolution. In practise, we found that both strategies lead to similar results during the self-supervised training. During the second training stage,  the weights in the training loss \eqref{eq:multiscale-loss} are instead set to $\gamma_{1}=0.08, \gamma_{2}=0.08, \gamma_{3}=0.02, \gamma_{4}=0.02$, which also ensures that the loss terms of all pyramid levels have the same magnitude. 

During the first training stage on uniquely the self-supervised data, we train for 135.000 iterations, with batch size of 15. The learning rate is initially equal to $10^{-4}$, and halved after 80.000 and 108.000 iterations. Note that during this first training stage, the feature back-bone is frozen, but further finetuned during the second training stage. 
While finetuning on the composition of MegaDepth and the synthetic dataset, the batch size is reduced to 10 and we further train for 195.000 iterations. The initial learning rate is fixed to $5.10^{-5}$ and halved after 120.000 and 180.000 iterations. The feature back-bone is also finetuned according to the same schedule, but with an initial learning rate of $10^{-5}$.
For the GOCor modules~\cite{GOCor}, we train with 3 local and global optimization iterations. 

Our system is implemented using Pytorch~\cite{pytorch} and our networks are trained using Adam optimizer~\cite{adam} with weight decay of $0.0004$.

\section{Experimental setup and datasets}
\label{sec-sup:details-evaluation}

In this section, we first provide details about the evaluation datasets and metrics. We then explain the experimental set-up in more depth. 

\subsection{Evaluation metrics}

\parsection{AEPE} AEPE is defined as the Euclidean distance between estimated and ground truth flow fields, averaged over all valid pixels of the reference image.

\parsection{PCK} The Percentage of Correct Keypoints (PCK) is computed as the percentage of correspondences $\mathbf{\tilde{x}}_{j}$ with an Euclidean distance error $\left \| \mathbf{\tilde{x}}_{j} - \mathbf{x}_{j}\right \| \leq  T$, w.r.t.\ to the ground truth $\mathbf{x}_{j}$, that is smaller than a threshold $T$.

\parsection{F1} F1 designates the percentage of outliers averaged over all valid pixels of the dataset~\cite{Geiger2013}. They are defined as follows, where $Y$ indicates the ground-truth flow field and $\hat{Y}$ the estimated flow by the network.
\begin{equation}
    F1 = \frac{    \left \| Y-\hat{Y} \right \| > 3  \textrm{ and }   \frac{\left \| Y-\hat{Y}  \right \|}{\left \| Y \right \|} > 0.05 } {\textrm{\#valid pixels}}
\end{equation}

\parsection{Sparsification Errors} Sparsification plots measure how well the estimated uncertainties fit the true errors. The pixels of a flow field are sorted according to their corresponding uncertainty, in descending order. An increasing percentage of the pixels is subsequently removed, and the AEPE or PCK of the remaining pixels is calculated. We refer to these curves as Sparsification. As reference, we also compute the Oracle, which represents the AEPE or PCK calculated when the pixels are ranked according to the true error, computed with the ground-truth flow field. Ideally, if the estimated uncertainty is a good representation of the underlying error distribution, the Sparsification should be close to the Oracle. To compare methods, since each approach results in a different oracle, we use the Area Under the Sparsification Error Curve (AUSE), where the Sparsification Error is defined as the difference between the sparsification and its oracle. We compute the sparsification error curve on each image pair, and normalize it to $[0,1]$. The final error curve is the average over all image pairs of the dataset.

\parsection{mAP} For the task of pose estimation, we use mAP as the evaluation metric, following~\cite{OANet}. The absolute rotation error $\left | R_{err}  \right |$ is computed as the absolute value of the rotation angle needed to align ground-truth rotation matrix $R$ with estimated rotation matrix $\hat{R}$, such as
\begin{equation}
    R_{err} = cos^{-1}\frac{Tr(R^{-1}\hat{R}) -1}{2} \;,
\end{equation} 
where operator $Tr$ denotes the trace of a matrix. The translation error $T_{err}$ is computed similarly, as the angle to align the ground-truth translation vector $T$ with the estimated translation vector $\hat{T}$. 
\begin{equation}
    T_{err} = cos^{-1}\frac{T \cdot \hat{T}}{\left \| T \right \|\left \| \hat{T} \right \|} \;,
\end{equation}
where $\cdot$ denotes the dot-product. The accuracy Acc-$\kappa$ for a threshold $\kappa$ is computed as the percentage of image pairs for which the maximum of $T_{err}$ and $\left | R_{err}  \right |$ is below this threshold. mAP is defined according to original implementation~\cite{OANet}, \ie mAP @5\textdegree\, is equal to Acc-5, mAP @10\textdegree\,  is the average of Acc-5 and Acc-10, while mAP @20\textdegree\,  is the average over Acc-5, Acc-10, Acc-15 and Acc-20. 

\subsection{Evaluation datasets and set-up}
\label{details-eval-data}

\parsection{MegaDepth} The MegaDepth dataset depicts real scenes with extreme viewpoint changes. No real ground-truth correspondences are available, so we use the result of SfM reconstructions to obtain sparse ground-truth correspondences. We follow the same procedure and test images than~\cite{RANSAC-flow}. More precisely, we randomly sample 1600 pairs of images that shared more than 30 points. The test pais are from different scenes than the ones we used for training and validation. We use 3D points from SfM reconstructions and project them onto the pairs of matching images to obtain correspondences. It results in approximately 367K correspondences. During evaluation, following~\cite{RANSAC-flow}, all the images are resized to have minimum dimension 480 pixels.

\parsection{RobotCar} In RobotCar, we used the correspondences originally introduced by~\cite{RobotCarDatasetIJRR}. During evaluation, following~\cite{RANSAC-flow}, all the images are resized to have minimum dimension 480 pixels.

\parsection{ETH3D}  The Multi-view dataset ETH3D~\cite{ETH3d} contains 10 image sequences at $480 \times 752$ or $514 \times 955$ resolution, depicting indoor and outdoor scenes. They result from the movement of a camera completely unconstrained, used for benchmarking 3D reconstruction. 
The authors additionally provide a set of sparse geometrically consistent image correspondences (generated by~\cite{SchonbergerF16}) that have been optimized over the entire image sequence using the reprojection error. We sample image pairs from each sequence at different intervals to analyze varying magnitude of geometric transformations, and use the provided points as sparse ground truth correspondences. This results in about 500 image pairs in total for each selected interval, or 600K to 1000K correspondences. Note that, in this work, we computed the PCK over the whole dataset per interval, to be consistent with RANSAC-Flow.  This metric is different than the one originally used by~\cite{GLUNet, GOCor} for ETH3D, where the PCK was calculated per image instead. 

\parsection{KITTI} The KITTI dataset~\cite{Geiger2013} is composed of real road sequences captured by a car-mounted stereo camera rig. The KITTI benchmark is targeted for autonomous driving applications and its semi-dense ground truth is collected using LIDAR. The 2012 set only consists of static scenes while the 2015 set is extended to dynamic scenes via human annotations.  The later contains  large motion, severe illumination changes, and occlusions. 

\parsection{YFCC100M} The YFCC100M dataset represents touristic landmark images. The ground-truth poses were created by generating 3D reconstructions from a subset of the collections~\cite{heinly2015_reconstructing_the_world}. 
Since our network PDC-Net outputs flow fields at a quarter of the images original resolution, which are then usually up-sampled, for pose estimation, we directly select matches at the outputted resolution and further scale the correspondences to original resolution. From the estimated dense flow field, we identify the accurate correspondences by thresholding the predicted confidence map $P_R$ (Sec \ref{sec:geometric-inference} of the main paper, and Sec.~\ref{sec-sup:proba}). Specifically, we select correspondences for which the corresponding confidence level at $R=1$ is superior to 0.1, such as $P_{R=1} > 0.1$. 
We then use the selected matches to estimate an essential matrix with RANSAC~\cite{Fishchler} and 5-pt Nister algorithm~\cite{TPAMI.2004.17}, relying on OpenCV’s 'findEssentialMat' with an inlier threshold of 1 pixel divided by the focal length. Rotation matrix $\hat{R}$ and translation vector $\hat{T}$ are finally computed from the estimated essential matrix, using OpenCV’s 'recoverPose'. The original images are resized to have a minimum dimension of 480, similar to~\cite{RANSAC-flow}, and the intrinsic camera parameters are modified accordingly.

\begin{table*}[t]
\centering
\resizebox{\textwidth}{!}{%
\begin{tabular}{lccc|ccc|cc|cc|ccc} \toprule
& \multicolumn{3}{c}{\textbf{MegaDepth}} & \multicolumn{3}{c}{\textbf{RobotCar}} & \multicolumn{2}{c}{\textbf{KITTI-2012}} & \multicolumn{2}{c}{\textbf{KITTI-2015}} & \multicolumn{3}{c}{\textbf{YFCC100M}} \\ 
 
& PCK-1  & PCK-3  & PCK-5 & PCK-1  & PCK-3 & PCK-5  & AEPE  $\downarrow$            & F1   (\%)   $\downarrow$      & AEPE  $\downarrow$              & F1  (\%)  $\downarrow$ &  mAP @5\textdegree &  mAP @10\textdegree & mAP @20\textdegree \\ \midrule
DGC-Net~\cite{Melekhov2019} & 3.55 & 20.33 & 32.28 & 1.19 & 9.35 & 20.17 &  8.50  &   32.28 &  14.97   &      50.98  & 6.73 & 12.55 & 22.42 \\
GLU-Net~\cite{GLUNet} & 21.58 & 52.18 & 61.78 & 2.30 & 17.15 & 33.87 & 3.14 & 19.76 & 7.49 & 33.83 & 21.35 & 30.73 & 42.91 \\
GLU-Net-GOCor~\cite{GOCor} &  37.28 &  61.18  & 68.08 & 2.31 & 17.62 & 35.18 & 2.68 & 15.43 & \textbf{6.68} & 27.57 & 24.53 & 33.56 & 45.34 \\ 
GLU-Net-GOCor* & 41.36 & 62.37 & 67.23 & 2.07 & 15.57  & 30.86 & 2.95 & 14.05 & 7.14 & 25.02 & 24.55 & 32.55 & 43.31 \\
\textbf{PDC-Net} & 53.06 & 70.88 & 73.94 & \textbf{2.54} & \textbf{18.85} & \textbf{36.24} & \textbf{2.44} & \textbf{11.09} & 6.82 & \textbf{21.79} & 47.40 & 56.46 & 65.33 \\ 
\textbf{PDC-Net (MS)} &  \textbf{56.49} & \textbf{76.65} & \textbf{80.18} & 2.53  & 18.68  & 36.03 &  - & -  & - & - &  \textbf{53.80} & \textbf{63.94} & \textbf{73.86} \\
\bottomrule
\end{tabular}%
}\vspace{1mm}\caption{Results on multiple geometric and optical flow datasets as well as for pose estimation on the YFCC100M dataset. All methods are trained on purely self-supervised data. 
}\vspace{-3mm}
\label{tab:synthetic-data}
\end{table*}

\begin{figure*}
\centering%
\includegraphics[width=0.99\textwidth, trim={0 1cm 0 2cm}]{ 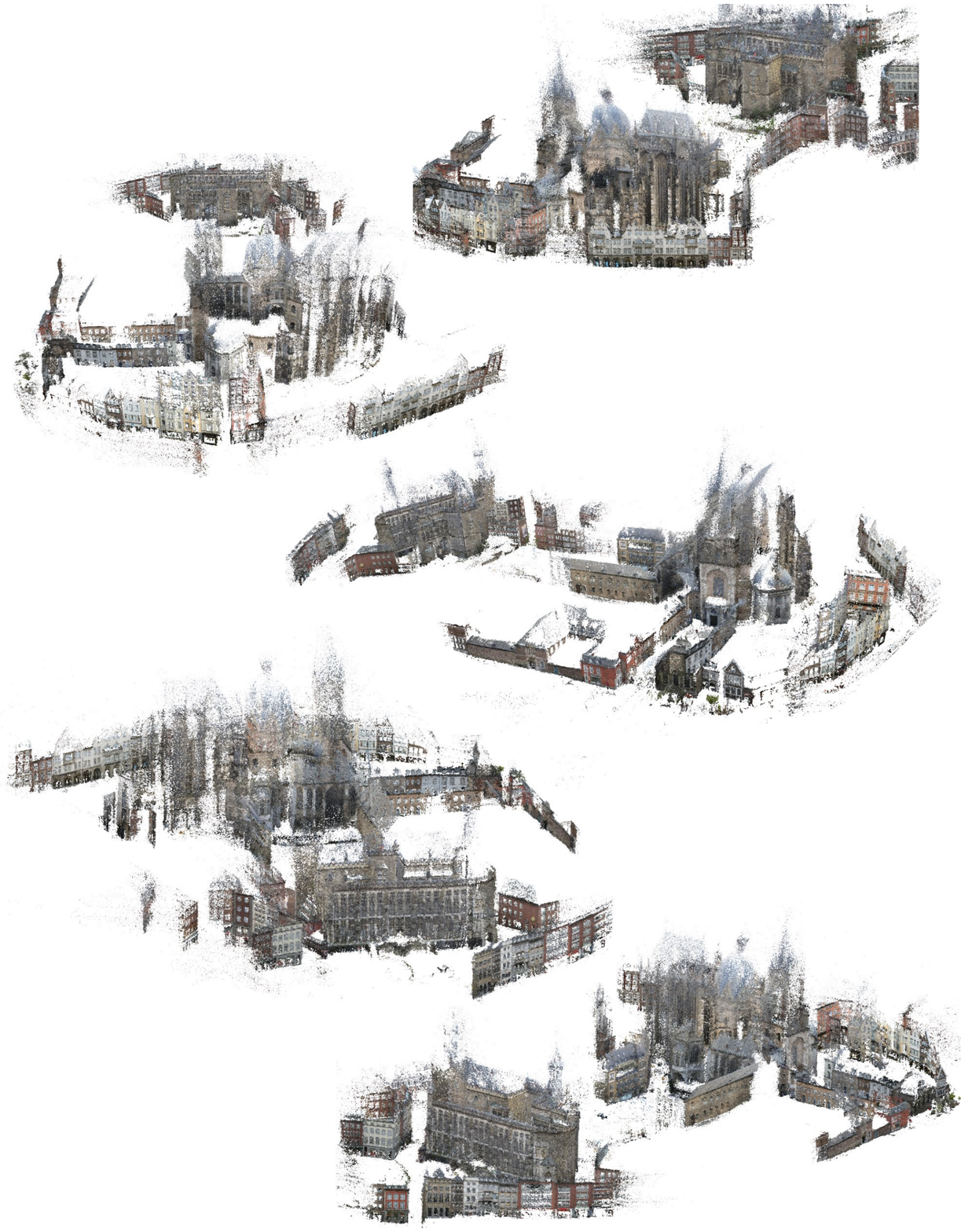}
\caption{Visualization of the 3D reconstruction of Aachen city. }
\label{fig:aachen-sup}
\end{figure*}

\parsection{3D reconstruction on Aachen dataset} We use the set-up of~\cite{SattlerMTTHSSOP18}, which provides a list of image pairs to match. We compute dense correspondences between each pair. We resize the images by keeping the same aspect ratio so that the minimum dimension is 600. We select matches for which the confidence probability $P_{R=1}$ is above 0.3, and feed them to COLMAP reconstruction pipeline~\cite{SchonbergerF16}. Again, we select matches at a quarter of the image resolution and scale the matches to original resolution. Following Fig.~\ref{fig:aachen} of the main paper, additional qualitative representations of the resulting 3D reconstruction are shown in Fig.~\ref{fig:aachen-sup}.

On all datasets, we use our multi-stage strategy (Sec. \ref{sec:geometric-inference} of the main paper), except for the KITTI datasets. Indeed, on optical flow data, which shows limited displacements and appearance variation, multi-stage strategy does not bring any improvements, and solely increases the run-time. For evaluation, we use 3 and 7 steepest descent iterations in the global and local GOCor modules~\cite{GOCor} respectively. 
We reported results of RANSAC-Flow~\cite{RANSAC-flow} using MOCO features, which gave the best results overall. %

\section{Detailed results}
\label{sec-sup:results}

In this section, we first provide detailed results on uncertainty estimation. Subsequently, we present results of our approach after solely the first training stage, \ie on uniquely self-supervised data. 
Finally, we present extensive qualitative results and comparisons.

\subsection{Detailed results on uncertainty estimation}

\begin{figure}[b]
\centering
\vspace{-6mm}\newcommand{\wid}{0.23\textwidth}
\subfloat{\includegraphics[width=\wid]{ 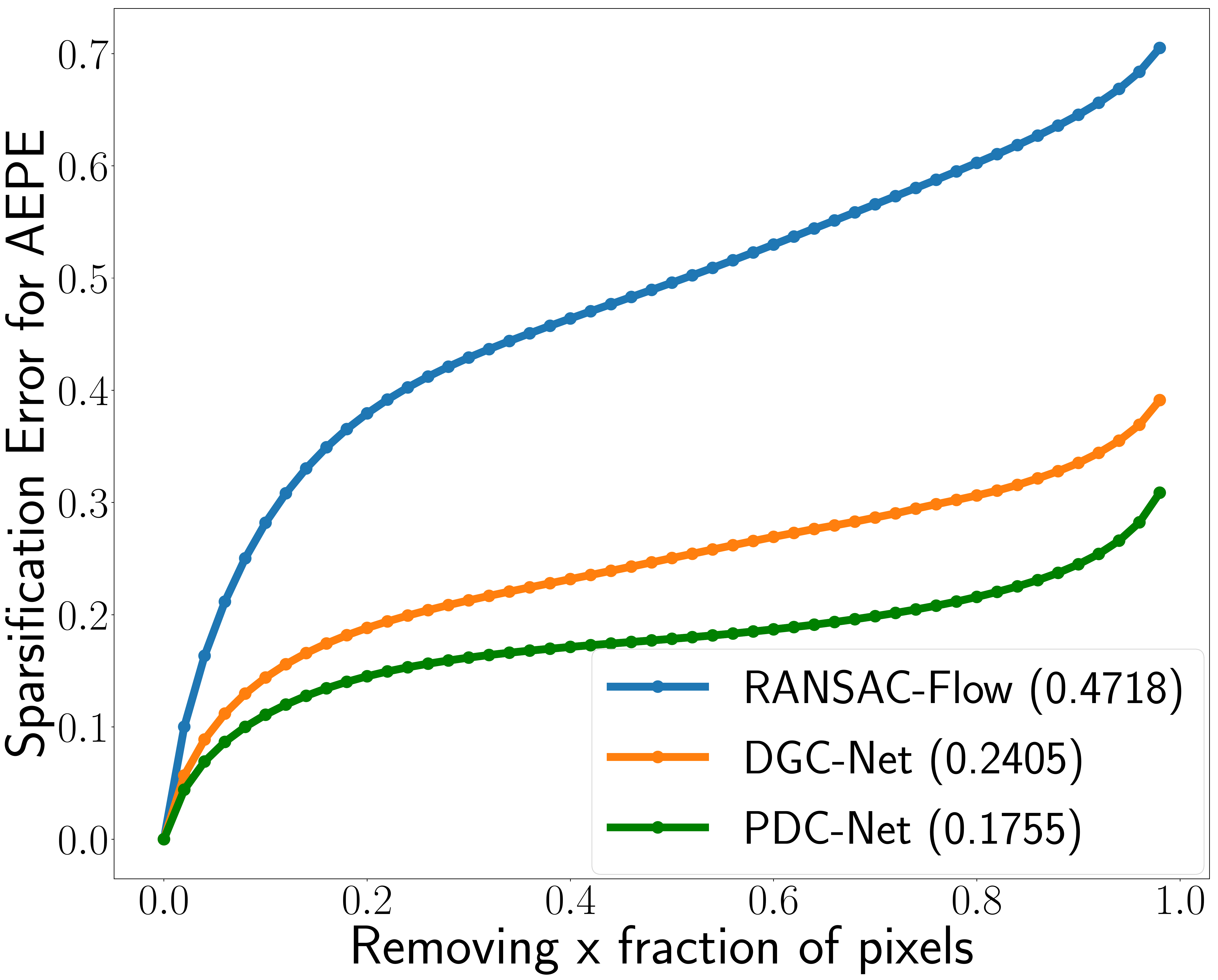}}~%
\subfloat{\includegraphics[width=\wid]{ 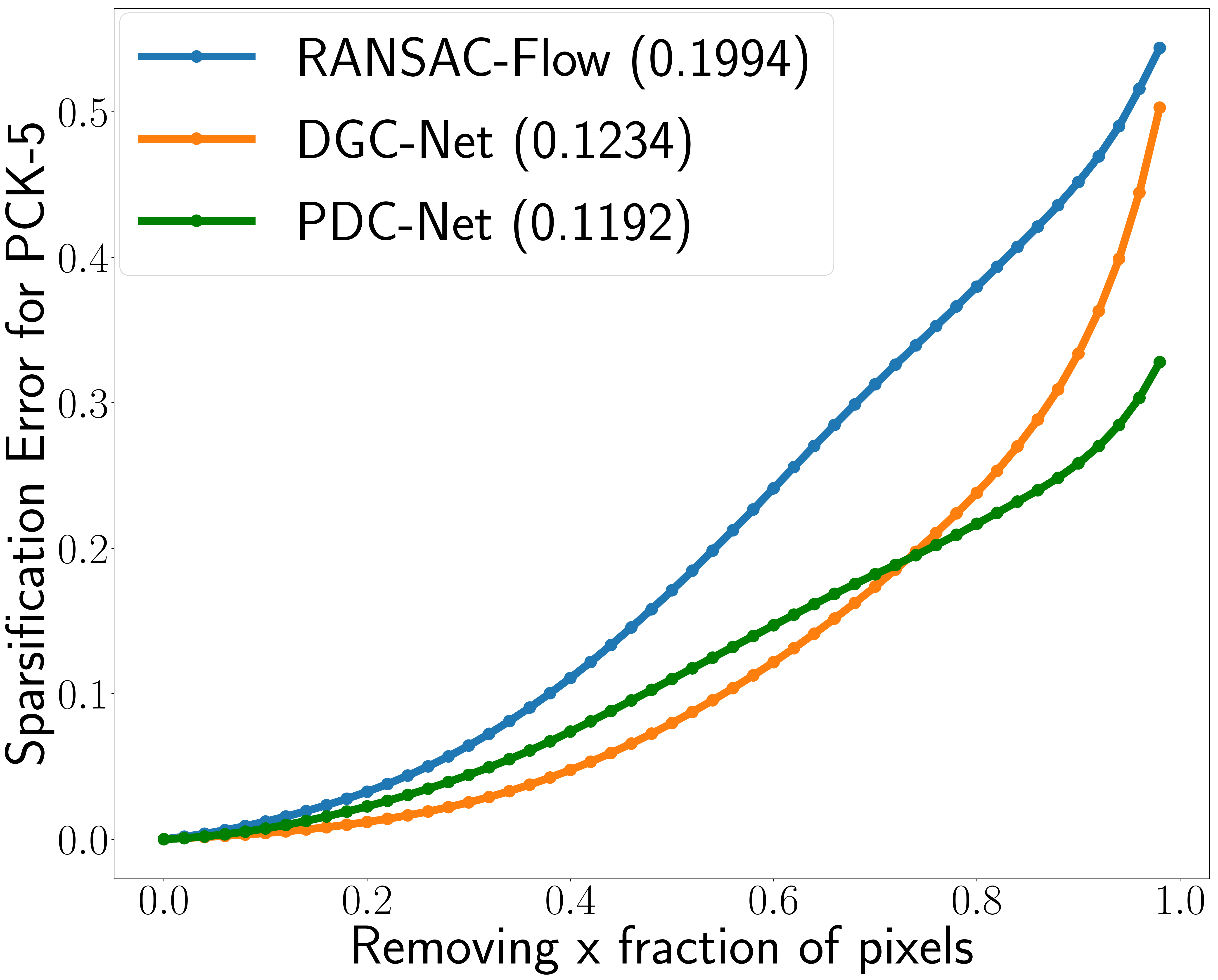}}~%
\caption{Sparsification Error plots for AEPE (left) and PCK-5 (right) on RobotCar.  Smaller AUSE (in parenthesis) is better. }
\label{fig:sparsification-robotcar}
\end{figure}

Here, we present sparsification errors curves, computed on the RobotCar dataset. As in the main paper, Sec. \ref{subsec:uncertainty-est}, we compare our probabilistic approach PDC-Net, to dense geometric methods providing a confidence estimation, DGC-Net~\cite{Melekhov2019} and RANSAC-Flow~\cite{RANSAC-flow}. Fig.~\ref{fig:sparsification-robotcar} depicts the sparsification error curves on RobotCar. As on MegaDepth, our approach PDC-Net estimates uncertainty map which better fit the underlying errors.

\begin{table*}[t]
\centering
\resizebox{0.99\textwidth}{!}{%
\begin{tabular}{lccc|ccc|cc} 
\toprule
& \multicolumn{3}{c}{\textbf{KITTI-2015}} & \multicolumn{3}{c}{\textbf{MegaDepth}} & \multicolumn{2}{c}{\textbf{YFCC100M}}\\
  & EPE  & F1 (\%)  & AUSE & PCK-1 (\%)  & PCK-5 (\%) & AUSE & mAP @5\textdegree & mAP @10\textdegree  \\ \midrule
Uncertainty not propagated between levels & 6.76 & \textbf{31.84} & 0.212 & 29.9 & 65.13 & 0.213 & 31.50 & 42.19  \\
PDC-Net-s  &  \textbf{6.66} &  32.32   &  \textbf{0.205}   &  \textbf{32.51}   &  \textbf{66.50}   &  \textbf{0.197}   &  \textbf{33.77}   &  \textbf{45.17}    \\
\midrule
$M=2$; $0.0 < \sigma^2_1 < 1.0$, $2.0 < \sigma^2_2 < \infty$ & 6.69 & 32.58 & 0.\textbf{181} & 32.47 &  65.45 & 0.205 & 30.50 & 40.75 \\

$M=2$; $\sigma^2_1=1.0$, $2.0 < \sigma^2_2 < \infty$ (PDC-Net-s) & 6.66 &  32.32   &  0.205   &  \textbf{32.51}   &  66.50   &  \textbf{0.197}   &  \textbf{33.77}   &  \textbf{45.17} \\

$M=2$; $\sigma^2_1=1.0$, $2.0 < \sigma^2_2 < \beta_2^+=s^2$ &  \textbf{6.61} & \textbf{31.67} & 0.208 & 31.83 & \textbf{66.52} & 0.204 & 33.05 & 44.48 \\

\midrule
$M=2$; $\sigma^2_1=1.0$, $2.0 < \sigma^2_2 < \beta_2^+=s^2$ &  6.61 & 31.67 & \textbf{0.208} & 31.83 & \textbf{66.52} & \textbf{0.204} & 33.05 & 44.48 \\
$M=3$; $\sigma^2_1=1.0$, $2.0 < \sigma^2_2 < \beta_2^+=s^2$, $\sigma^2_3 = s^2$ &  \textbf{6.41} & \textbf{30.54} & 0.212 & \textbf{31.89} & 66.10 & 0.214 & \textbf{34.90} & \textbf{45.86} \\
\bottomrule
\end{tabular}%
}\vspace{1mm}
\caption{Ablation study. For all methods, we model the flow as a constrained mixture of Laplace distributions (Sec. \ref{sec:constained-mixture} of the main paper), and we use our uncertainty prediction architecture (Sec. \ref{sec:uncertainty-arch} of the main paper). In the top part, we show the impact of propagating the uncertainty estimates in a multi-scale architecture. 
In the second part, we compare different parametrization of the constrained mixture (Sec. \ref{sec:constained-mixture} of the main paper, mostly equation \eqref{eq:constraint}). Here, $s$ refers to the image size used during training, \ie $s=256$. In the bottom part, we analyse the impact of the number of components $M$ used in the constrained mixture model. 
}\vspace{-1mm}
\label{tab:ablation-sup}
\end{table*}

\subsection{Results when trained on purely synthetic data}

For completeness, here, we show results of our approach PDC-Net, after only the first stage of training (described in Sec. \ref{subsec:arch} of the main paper and Sec.~\ref{sec-sup:training}), \ie after training on purely synthetically generated image warps, on which we overlaid moving objects and our flow perturbations (Sec. \ref{subsec:perturbed-data} of the main paper). For fair comparison, we compare to self-supervised approaches that also only rely on synthetic image warps, namely DGC-Net~\cite{Melekhov2019}, GLU-Net~\cite{GLUNet} and GLU-Net-GOCor~\cite{GLUNet, GOCor}. We also include our baseline, the non probabilistic model GLU-Net-GOCor* trained on the same data. Results on multiple datasets are presented in Tab.~\ref{tab:synthetic-data}. 
Our approach PDC-Net outperforms all other self-supervised methods, particularly in terms of accuracy (PCK and F1). Notably, our probabilistic modeling of the problem improves the learning, leading to large gains in flow estimation accuracy, as evidenced by comparing PDC-Net to non probabilistic baseline GLU-Net-GOCor*. Also note that for a \emph{single} forward pass, PDC-Net only increases inference time by $14.3\%$ over baseline GLU-Net-GOCor* for drastically better results. 

Moreover, for the non probabilistic methods, we computed the relative poses on YFCC100M using \emph{all} estimated dense correspondences and RANSAC. As previously stated, the poor results emphasize the necessity to infer a confidence prediction along with the dense flow prediction, in order to be able to use the estimated matches for down-stream tasks. 

\subsection{Qualitative results}

\begin{figure*}
\centering%
\includegraphics[width=0.95\textwidth]{ 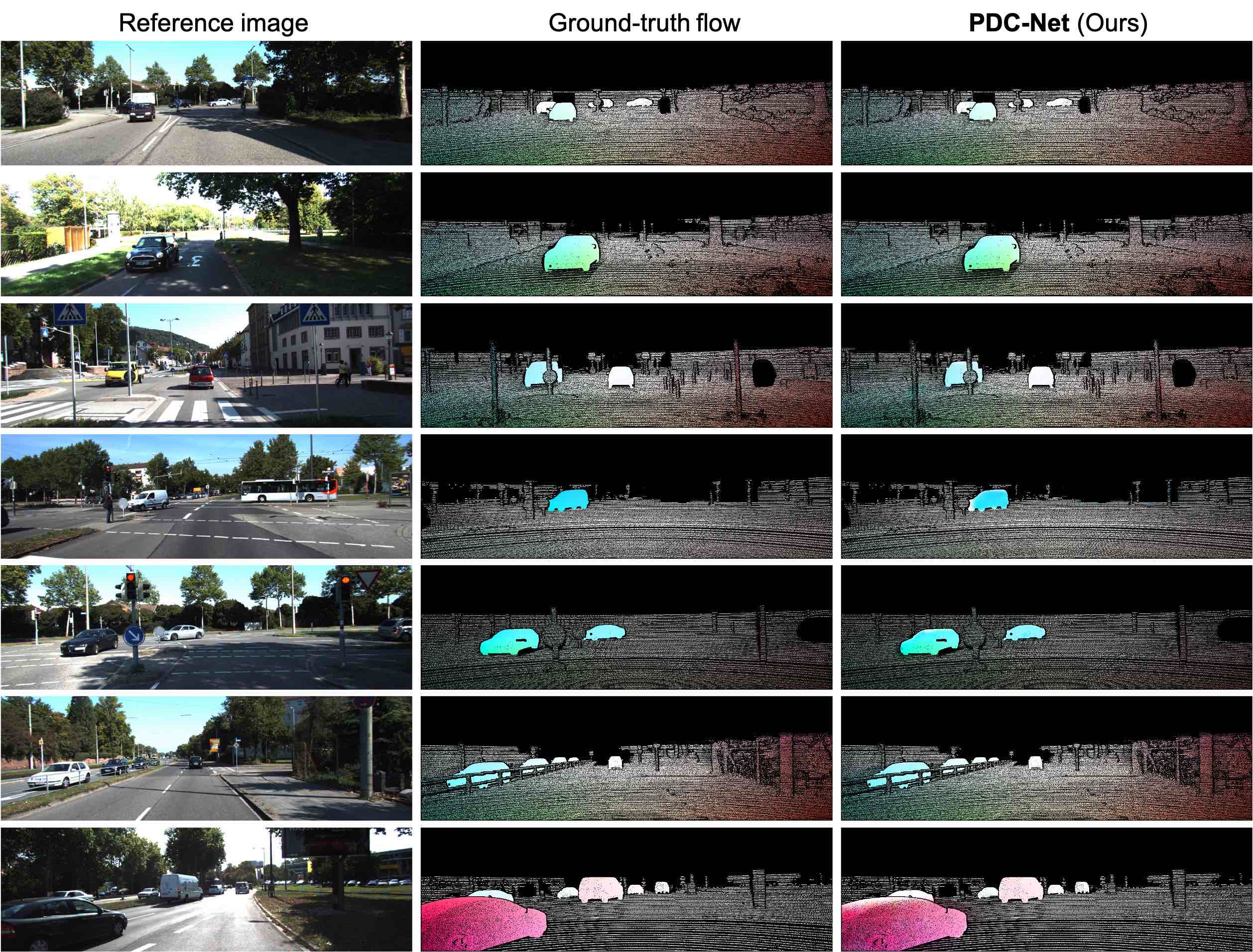}
\caption{Qualitative examples of our approach PDC-Net applied to images of KITTI-2015. We plot directly the estimated flow field for each image pair.}
\label{fig:kitti-qual}
\end{figure*}

Here, we first present qualitative results of our approach PDC-Net on the KITTI-2015 dataset in Fig.~\ref{fig:kitti-qual}. PDC-Net clearly identifies the independently moving objects, and does very well in static scenes with only a single moving object, which are particularly challenging since not represented in the training data. 

\begin{figure*}
\centering%
\includegraphics[width=0.99\textwidth]{ 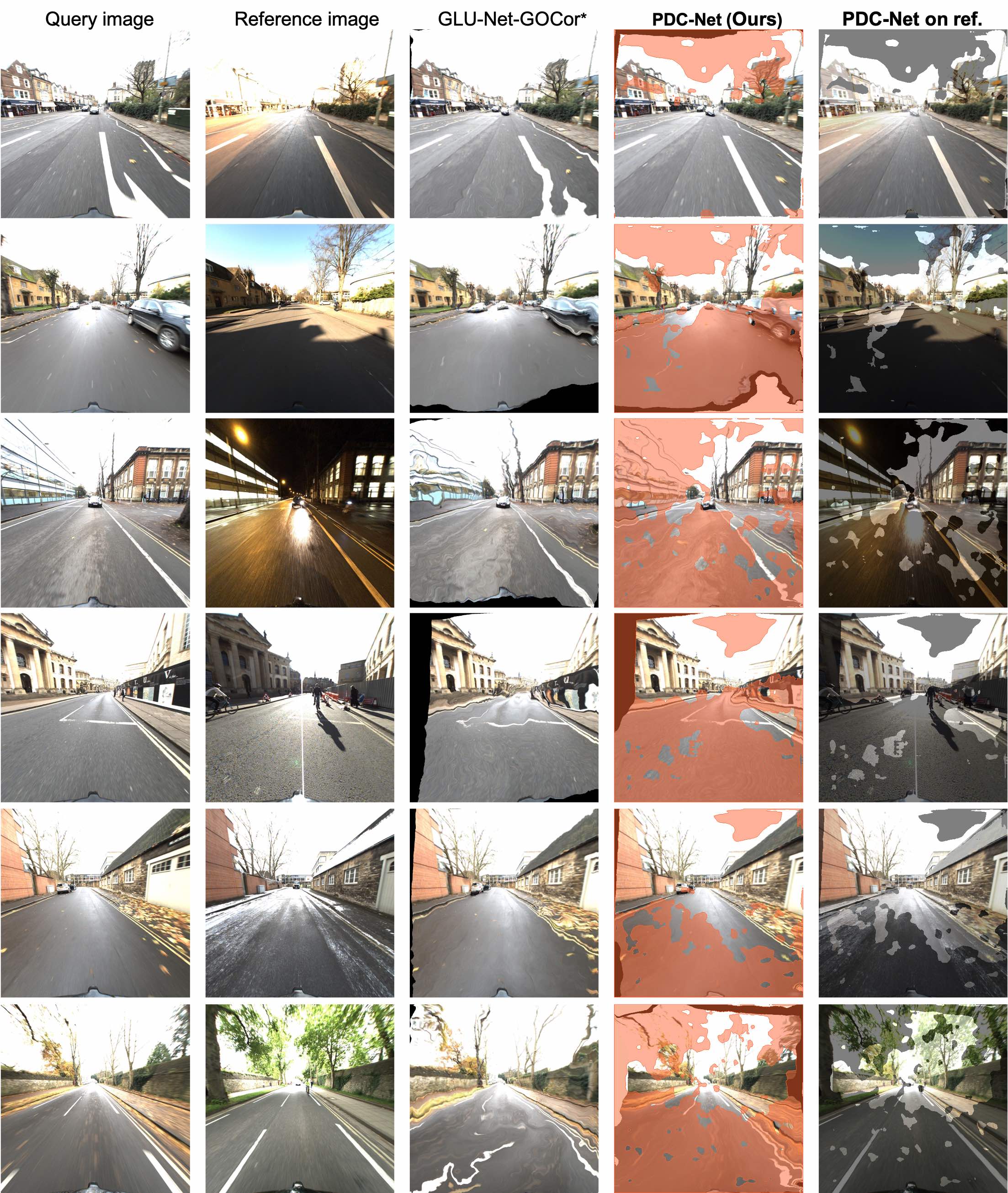}
\caption{Qualitative examples of our approach PDC-Net and corresponding non-probabilistic baseling GLU-Net-GOCor*, applied to images of the RobotCar dataset~\cite{RobotCar}. In the 3$^{rd}$ and 4$^{th}$ columns, we visualize the query images warped according to the flow fields estimated by the GLU-Net-GOCor* and PDC-Net respectively. PDC-Net also predicts a confidence map, according to which the regions represented in red, are unreliable or inaccurate matching regions. In the last column, we overlay the reference image with the warped query from PDC-Net, in the identified accurate matching regions (lighter color).}
\label{fig:robotcar-qual}
\end{figure*}

\begin{figure*}
\centering%
\vspace{-16mm}\includegraphics[width=0.93\textwidth]{ 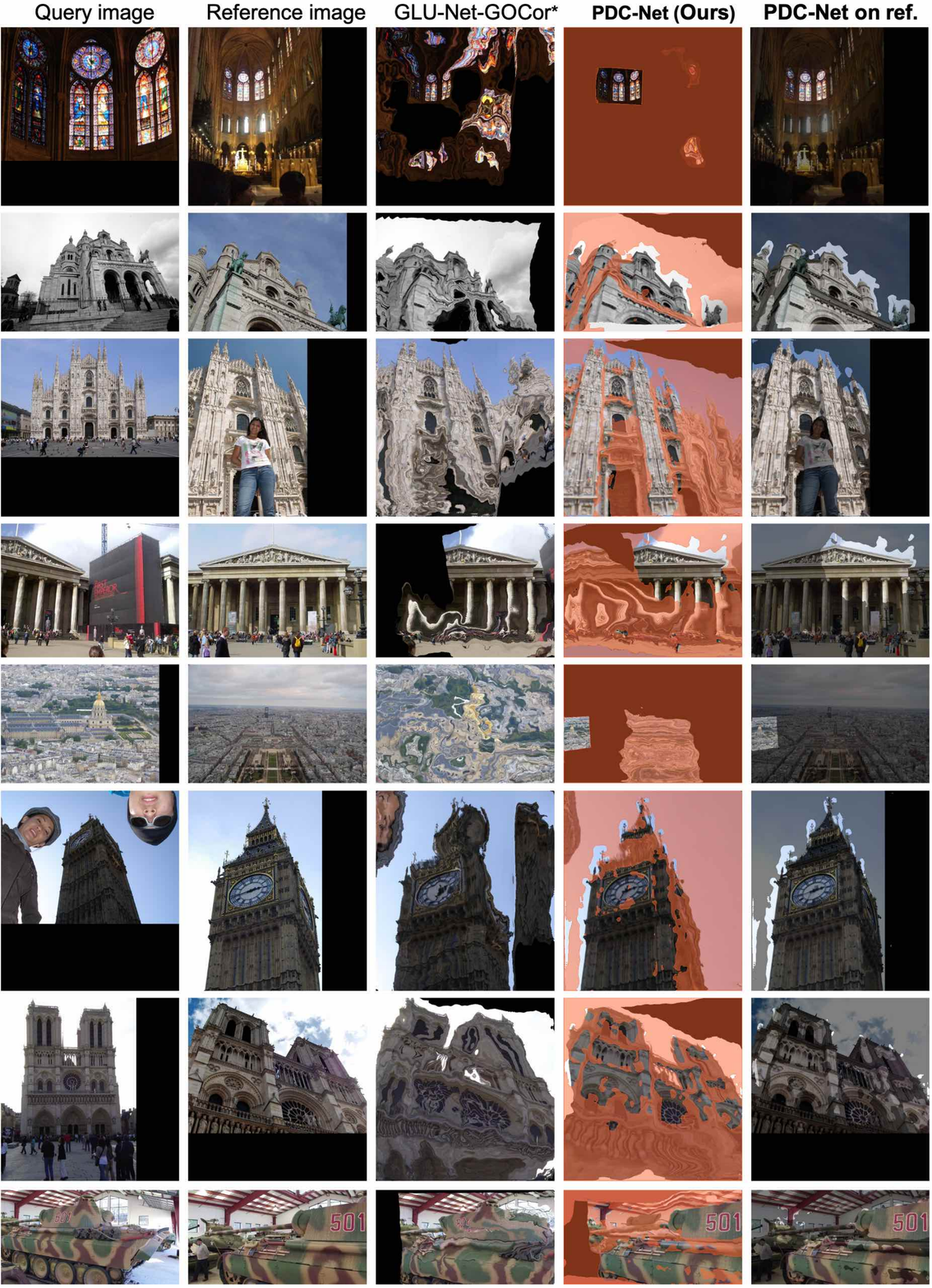}
\vspace{-2mm}\caption{Qualitative examples of our approach PDC-Net and corresponding non-probabilistic baseline GLU-Net-GOCor*, applied to images of the MegaDepth dataset~\cite{megadepth}. In the 3$^{rd}$ and 4$^{th}$ columns, we visualize the query images warped according to the flow fields estimated by the GLU-Net-GOCor* and PDC-Net respectively. PDC-Net also predicts a confidence map, according to which the regions represented in red, are unreliable or inaccurate matching regions. In the last column, we overlay the reference image with the warped query from PDC-Net, in the identified accurate matching regions (lighter color).}
\label{fig:mega-1}
\end{figure*}

\begin{figure*}
\centering%
\vspace{-15mm}\includegraphics[width=0.93\textwidth]{ 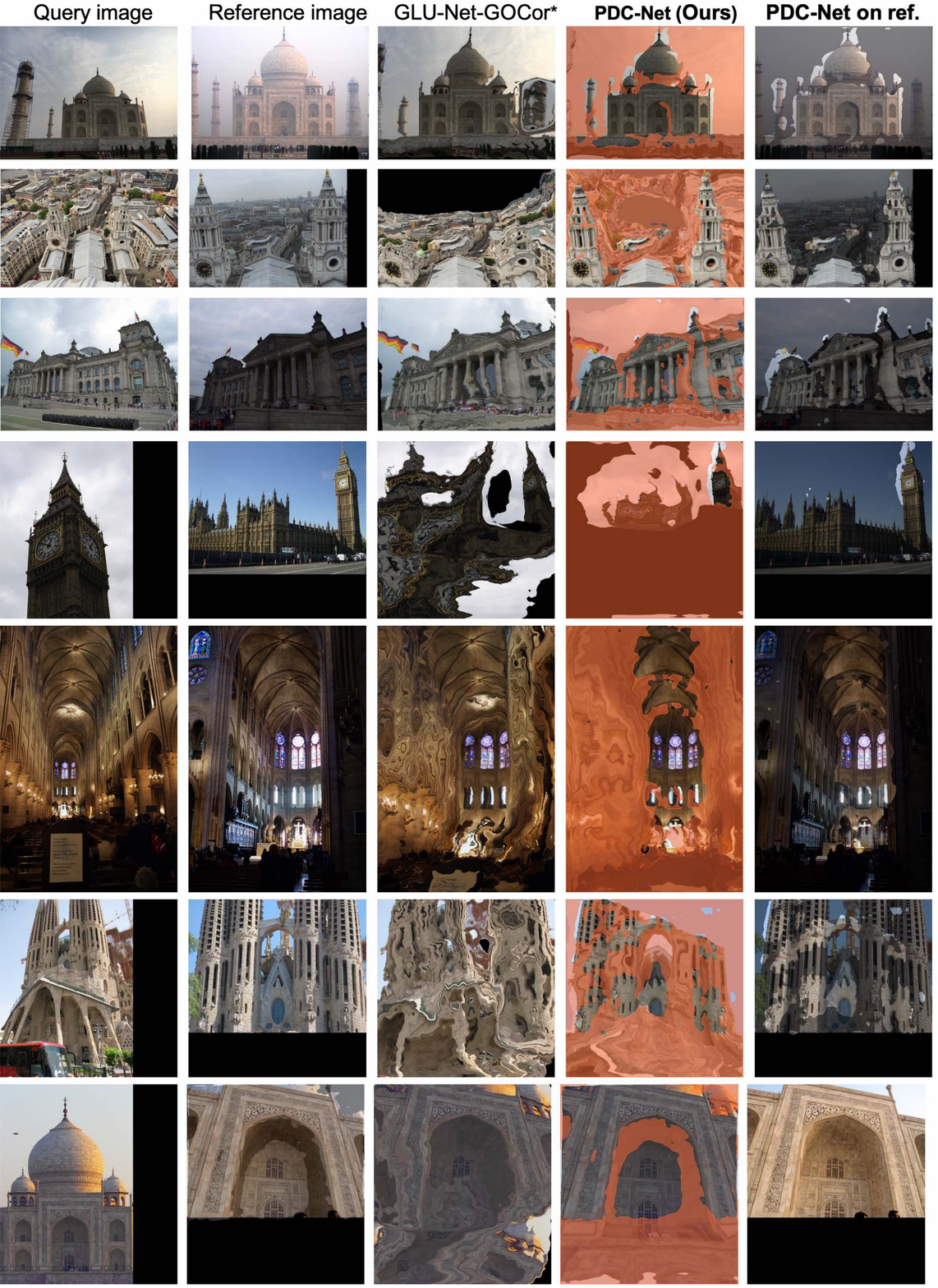}
\vspace{-5mm}\caption{Qualitative examples of our approach PDC-Net and corresponding non-probabilistic baseline GLU-Net-GOCor*, applied to images of the MegaDepth dataset~\cite{megadepth}. In the 3$^{rd}$ and 4$^{th}$ columns, we visualize the query images warped according to the flow fields estimated by the GLU-Net-GOCor* and PDC-Net respectively. PDC-Net also predicts a confidence map, according to which the regions represented in red, are unreliable or inaccurate matching regions. In the last column, we overlay the reference image with the warped query from PDC-Net, in the identified accurate matching regions (lighter color).}
\label{fig:mega-2}
\end{figure*}

\begin{figure*}
\centering%
\vspace{-13mm}\includegraphics[width=0.93\textwidth]{ 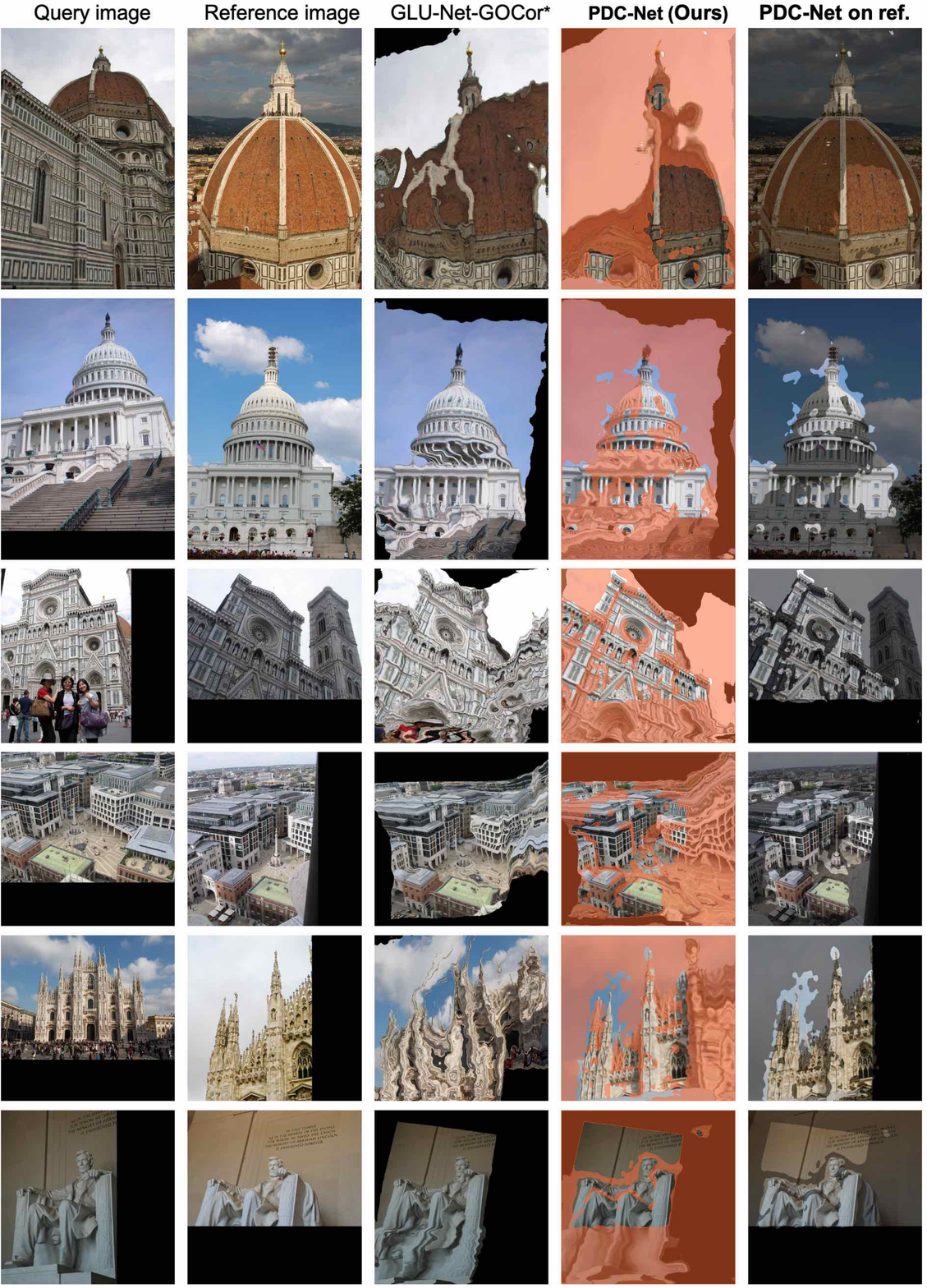}
\vspace{-5mm}\caption{Qualitative examples of our approach PDC-Net and corresponding non-probabilistic baseline GLU-Net-GOCor*, applied to images of the MegaDepth dataset~\cite{megadepth}. In the 3$^{rd}$ and 4$^{th}$ columns, we visualize the query images warped according to the flow fields estimated by the GLU-Net-GOCor* and PDC-Net respectively. PDC-Net also predicts a confidence map, according to which the regions represented in red, are unreliable or inaccurate matching regions. In the last column, we overlay the reference image with the warped query from PDC-Net, in the identified accurate matching regions (lighter color).}
\label{fig:mega-3}
\end{figure*}

In Fig.~\ref{fig:robotcar-qual} and Fig.~\ref{fig:mega-1},~\ref{fig:mega-2},~\ref{fig:mega-3}, we qualitatively compare our approach PDC-Net to the baseline GLU-Net-GOCor* on images of the RobotCar and the Megadepth datasets respectively. We additionally show the performance of our uncertainty estimation on these examples. By overlaying the warped query image with the reference image at the locations of the identified accurate matches, we observe that our method produces \emph{highly precise correspondences}. Our uncertainty estimates successfully identify accurate flow regions and also correctly exclude in most cases homogeneous and sky regions. These examples show the benefit of confidence estimation for high quality image alignment, useful \eg in multi-frame super resolution~\cite{WronskiGEKKLLM19}. Texture or style transfer (\eg for AR) also largely benefit from it. 

\begin{figure*}
\centering%
\vspace{-5mm}\includegraphics[width=0.99\textwidth]{ 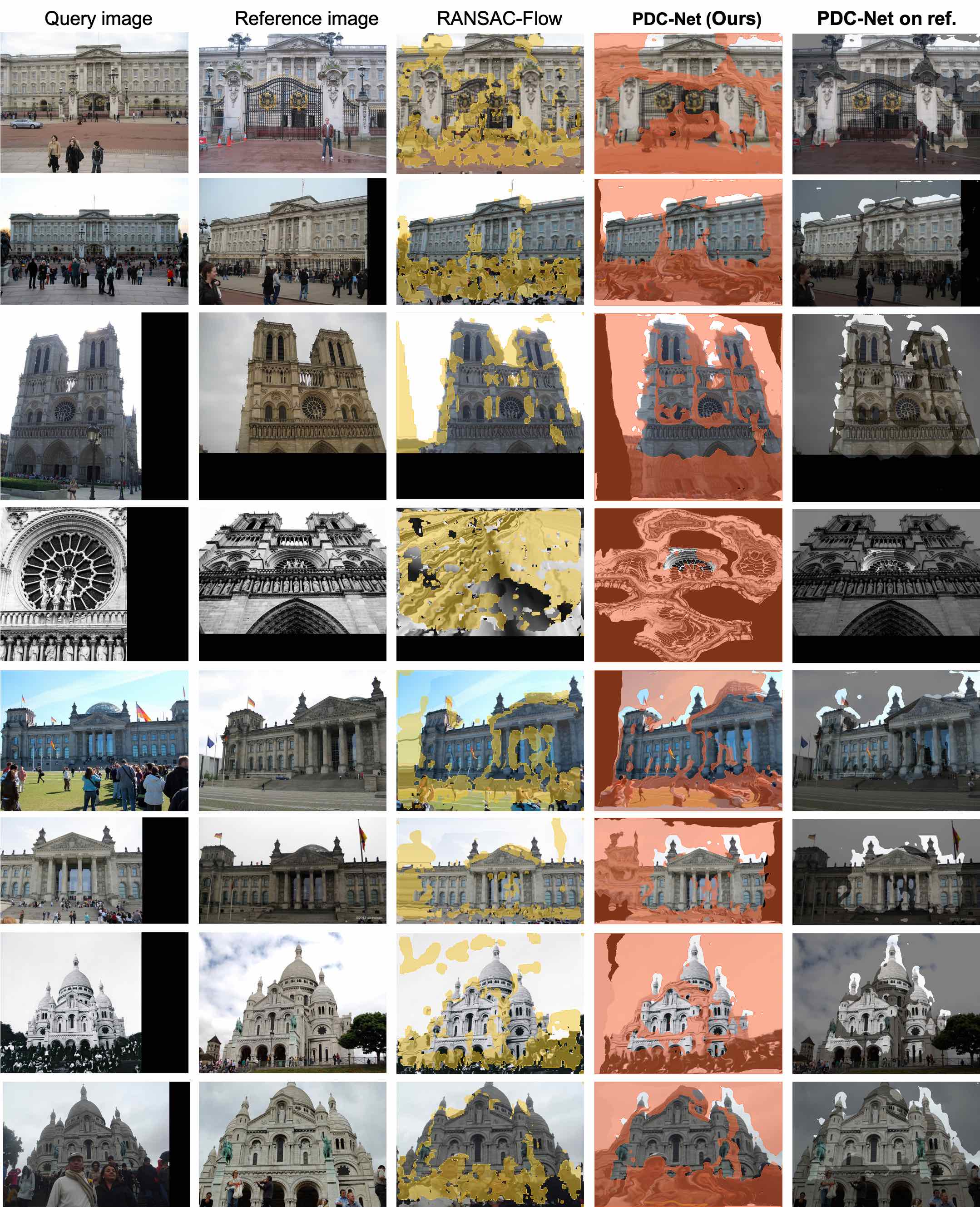}
\caption{Visual comparison of RANSAC-Flow and our approach PDC-Net on image pairs of the YFCC100M dataset~\cite{YFCC}. In the 3$^{rd}$ and 4$^{th}$ columns, we visualize the query images warped according to the flow fields estimated by the RANSAC-Flow and PDC-Net respectively. Both networks also predict a confidence map, according to which the regions represented in respectively yellow and red, are unreliable or inaccurate matching regions. In the last column, we overlay the reference image with the warped query from PDC-Net, in the identified accurate matching regions (lighter color). }
\label{fig:YCCM-qual}
\end{figure*}

In Fig.~\ref{fig:YCCM-qual}, we visually compare the estimated confidence maps of RANSAC-Flow~\cite{RANSAC-flow} and our approach PDC-Net on the YFCC100M dataset. Our confidence maps can accurately \textit{segment} the object from the background (sky). On the other hand, RANSAC-Flow predicts confidence maps, which do not exclude unreliable matching regions, such as the sky. Using these regions for pose estimation for example, would result in a drastic drop in performance, as evidenced in Tab. \ref{tab:YCCM} of the main paper. 
Note also the ability of our predicted confidence map to identify \emph{small accurate flow regions}, even in a dominantly failing flow field. This is the case in the fourth example from the top in Fig.~\ref{fig:YCCM-qual}.

\section{Detailed ablation study}
\label{sec-sup:ablation}

\begin{figure}[b]
\centering
\vspace{-3mm}\newcommand{\wid}{0.49\textwidth}
\includegraphics[width=\wid]{ 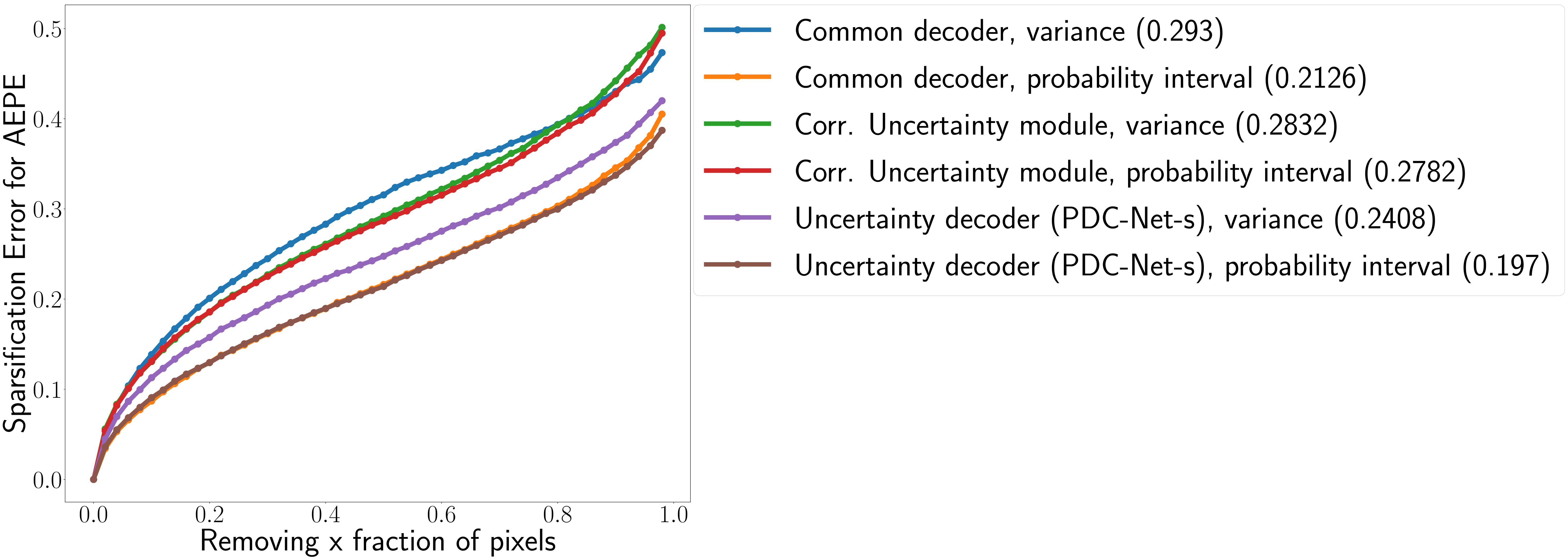} \\
\includegraphics[width=\wid]{ 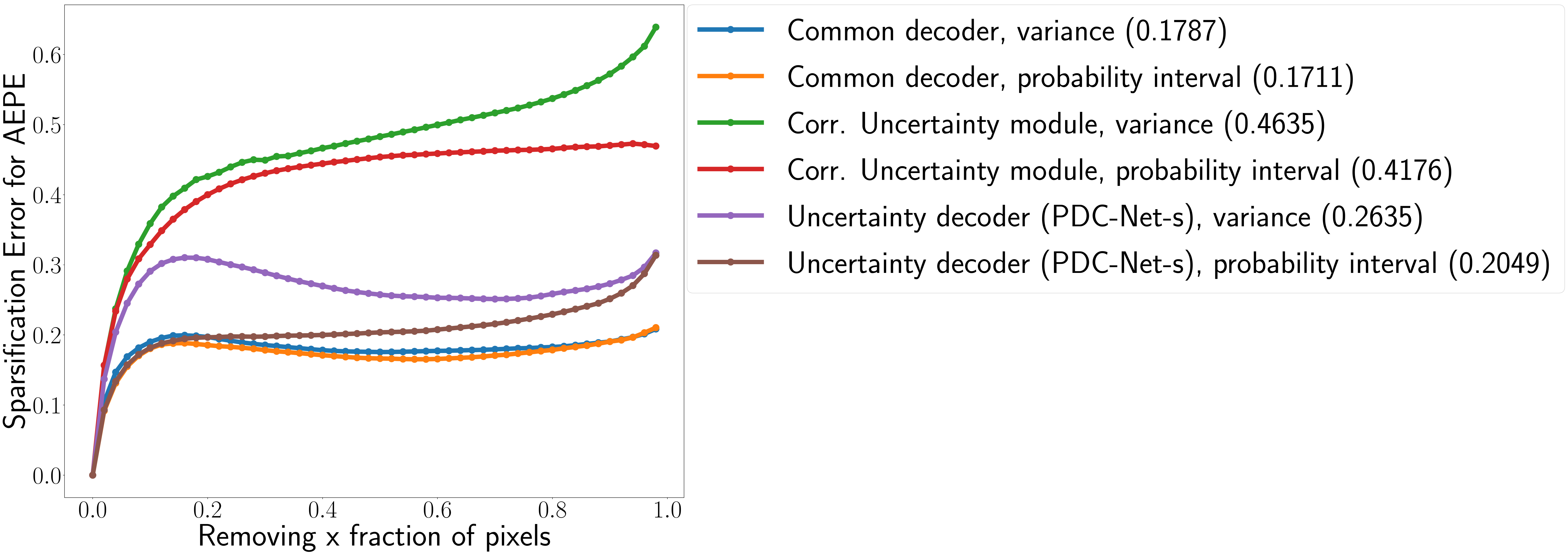} \\
\caption{Sparsification Error plots for AEPE on MegaDepth (top) and KITTI-2015 (bottom), for different uncertainty decoder architecture, and using either the variance $V$ or our probability interval $P_{R=1}$ as confidence measure.  Smaller AUSE (in parenthesis) is better. All networks are modelled with a constrained mixture of Laplace, and trained only on the first stage (Sec. \ref{subsec:arch} of the main paper).}
\label{fig:sparsification-ab}
\end{figure}

Finally, we provide detailed ablative experiments. As in Sec. \ref{subsec:ablation-study} of the main paper, we use BaseNet as base network to create the probabilistic models, as described in Sec.~\ref{sec-sub:arch-details}. Similarly, all networks are trained on solely the first training stage. 

\parsection{Uncertainty architecture, sparsification plots} For completeness, we present the full sparsification error plots for different uncertainty prediction architectures in Fig.~\ref{fig:sparsification-ab}. They correspond to Tab. \ref{tab:ablation}, middle part of the main paper. 

\parsection{Confidence value, variance against probability of interval} We here compare using the variance of the constrained mixture probability density $V = \sum_{m=1}^{M} \alpha_m \sigma^2_m$, or the probability of the confidence interval $P_R$ (Sec. \ref{sec:geometric-inference} of the main paper), as a final pixel-wise confidence value, associated with the predicted flow field. In Fig.~\ref{fig:sparsification-ab}, for each of the compared uncertainty prediction architecture, we further compute the sparsification error plots, using either the inverse of the variance $1/V$, or the probability $P_{R=1}$, as confidence measure. On both MegaDepth and KITTI-2015, the probability of the confidence interval $P_{R=1}$ appears as a better performing measure of the uncertainty. 

\parsection{Propagation of uncertainty components} In a multi-scale network architecture, the predicted uncertainty parameters of the mixture at a particular network level can be further propagated to the next level. We use this strategy, by feeding the predicted uncertainty components of the previous level to the flow decoder and to the uncertainty predictor of the current level. In Tab. \ref{tab:ablation-sup} top part, we show the impact of this uncertainty propagation. We compare our approach PDC-Net-s with multi-scale uncertainty propagation, to a network where uncertainty estimation at each level is done independently of the previous one. For all presented datasets and metrics, propagating the uncertainty predictions boosts the performance of the final network. Only the F1 metric on KITTI-2015 is slightly worst. 

\parsection{Constrained mixture parametrization} In Tab. \ref{tab:ablation-sup}, middle part, we then compare different parametrization for the constrained mixture of Laplace distributions. In our final network, we fixed the first component $\sigma^2_1$ of the mixture, as $\sigma^2_1=\beta_1^- =\beta_1^+ = 1.0$. Firstly, we compare with a version where the first component is constrained instead as $\beta_1^- = 0.0 < \sigma^2_1 < \beta_1^+=1.0$. Both networks obtain similar flow performance results on KITTI-2015 and MegaDepth. Only on optical flow dataset KITTI-2015, the alternative of $\sigma^2_1=\beta_1^- =\beta_1^+ = 1.0$ obtains a better AUSE, which is explained by the fact that KITTI-2015 shows ground-truth displacements with a much smaller magnitude than in the geometric matching datasets. When estimating the flow on KITTI, it thus results in a larger proportion of very small flow errors (lower EPE and higher PCK than on geometric matching data). As a result, on this dataset, being able to model very small error (with $\sigma^2_1 < 1$) is beneficial.  However, fixing $\sigma^2_1=1.0$  instead produces better AUSE on MegaDepth and it gives significantly better results for pose estimation on YFCC100M. As previously explained in Sec. \ref{subsec-sup:pdcnet}, fixing $\sigma^2_1=1$ enables for the network to equally focus on getting accurate correspondences (bounded by the fixed $\sigma^2_1$) and improving the inaccurate flow regions during training. It can be seen as introducing an additional prior constraint on the distribution. 

We then compare leaving the second component's higher bound unconstrained, as $2.0 < \sigma^2_2 < \infty$ (PDC-Net-s) to constraining it, as $2.0 < \sigma^2_2 < \beta_2^+ = s^2$, where s refers to the image size used during training.  All results are very similar, the fully constrained network obtains slightly better flow results but slightly worst uncertainty performance (AUSE). However, we found that constraining $\beta_2^+$ leads to a more stable training in practise, which is why we adopted the constrains  $\sigma^2_1 = \beta_1^- = \beta_1^+=1$, and $2.0 = \beta_2^- \leq \sigma^2_2 \leq \beta_2^+ = s^2$ for our final network.

\parsection{Number of components of the constrained mixture} Finally, we compare $M=2$ and $M=3$ Laplace components used in the constrained mixture in Tab. \ref{tab:ablation-sup}, bottom part. In the case of $M=3$, the first two components are set similarly to the case $M=2$, \ie as $\sigma^2_1=1.0$ and $2.0 < \sigma^2_2 < \beta_2^+=s^2$ where $\beta_2^+$ is fixed to the image size used during training (256 here). The third component is set as $\sigma^2_3 = \beta_3^+ = \beta_3^- = \beta_2^+$. The aim of this third component is to identify outliers (such as out-of-view pixels) more clearly. The 3 components approach obtains a better F1 value on KITTI-2015 and slightly better pose estimation results on YFCC100M. However, its performance on MegaDepth and in terms of pure uncertainty estimation (AUSE) slightly degrade. As a result, for simplicity we adopted the version with $M=2$ Laplace components. Note, however, that more components could easily be added.

\end{document}